\PassOptionsToPackage{table}{xcolor}
\documentclass{article}
\usepackage{iclr2027_conference,times}

\usepackage{xcolor}
\usepackage[utf8]{inputenc}
\usepackage[T1]{fontenc}
\usepackage{url}
\usepackage{xurl}
\usepackage{booktabs}
\usepackage{array}
\usepackage{amsmath}
\usepackage{amsfonts}
\usepackage{nicefrac}
\usepackage{microtype}
\usepackage{multirow}
\usepackage{graphicx}
\usepackage{float}
\usepackage{placeins}
\usepackage{hyperref}
\hypersetup{hidelinks}
\newcolumntype{R}[1]{>{\raggedleft\arraybackslash}p{#1}}

\newcommand{\appendixtablestyle}{%
  \small
  \setlength{\tabcolsep}{3pt}%
  \renewcommand{\arraystretch}{1.15}%
}

\title{\textsc{ResComEmb}: Effective and Efficient Multimodal Embedding via Residual Homogeneity Compression}

\author{
  Zijing Cai$^{1,*}$ \quad Yuzhe Wang$^{1,*}$ \quad
  Jingxian Zhu$^2$ \quad Fengbin Zhu$^{3,\dagger}$ \quad Richang Hong$^2$ \\
  $^1$University of Science and Technology of China \\
  $^2$Hefei University of Technology \\
  $^3$National University of Singapore
}

\iclrfinalcopy

\begin{document}

\maketitle
\begingroup
\renewcommand{\thefootnote}{\fnsymbol{footnote}}
\footnotetext[1]{Equal contribution.}
\footnotetext[2]{Corresponding author.}
\endgroup
\lhead{Preprint}

\begin{abstract}
Multimodal large language models (MLLMs) have shown strong potential for universal multimodal representation learning.
However, existing methods either compress each input into a single vector, limiting fine-grained expressiveness, or retain long sequences of visual-token vectors, incurring substantial storage and interaction costs. To resolve this trade-off, we propose \textsc{ResComEmb}, a trainable framework for effective and efficient universal multi-vector multimodal embedding. \textsc{ResComEmb} first encodes each input at native dynamic resolution into ordered global, intermediate, and fine-grained views.
After MLLM contextualization and embedding projection, a trainable Residual Homogeneity Compression (RHC) module reduces within-granularity redundancy and cross-granularity repetition under explicit visual token budgets.
Then, \textsc{ResComEmb} introduces a length-adaptive Bidirectional Late-Interaction Matching mechanism for robust query–document scoring, which averages the strongest token-level matches in each direction and combines the two scores using a weight based on how many valid tokens each side has.
Extensive experiments on MMEB, ViDoRe V1, and ViDoRe V2 show that ResComEmb produces higher-quality universal multimodal embeddings than VLM2Vec-V2, and outperforms ColQwen2.5 in visual document retrieval using only 37.5\% of its full visual token budget, demonstrating a favorable effectiveness–efficiency trade-off.

\end{abstract}

\section{Introduction}
\label{sec:intro}

Multimodal embedding models project heterogeneous inputs into a shared semantic space, enabling downstream applications, including image classification~\citep{deng2009imagenet}, visual question answering~\citep{hu2018learning}, cross-modal retrieval~\citep{gordo2016deep}, and visual grounding~\citep{refcoco}. 
Contrastive vision--language encoders~\citep{radford2021learning,zhai2023sigmoid},
learn cross-modal alignment from large-scale paired data with separate modality towers. Although compact and efficient, this modality-separated design struggles with interleaved image--text inputs and complex instructions~\citep{jiang2024e5v,zhang2024longclip}. 
Recent work therefore adapts multimodal large language models (MLLMs), such as Qwen2.5-VL~\citep{Qwen25-VL} and PaliGemma~\citep{beyer2024paligemma}, into universal multimodal embedding models, leveraging their pre-trained knowledge, multimodal understanding, and instruction-following capabilities~\citep{jiang2025vlmvec}.

MLLM-based embedding models represent visual inputs with a single vector~\citep{jiang2025vlmvec} or multiple vectors~\citep{ColPali,zhu2026mure}, creating a fidelity--efficiency trade-off. A single vector is compact but can limit the expression of fine-grained information in visually rich inputs~\citep{yao2022filip,thrush2022winoground}. Multiple vectors offer higher representational capacity by preserving local evidence, but increase storage and pairwise interaction costs~\citep{ColPali}. Existing approaches therefore pursue compact multi-vector representations through two routes. Token pruning and merging retain features aligned with the original visual sequence, but aggressive reduction sacrifices holistic context~\citep{visionzip2024,folder2025,dart2025}. Learnable-token methods instead condense visual features through a fixed number of query slots, efficiently capturing global semantics; under tight budgets, however, their fixed capacity restricts locality and input-specific details~\citep{cha2024honeybee,liu2024visual}. Neither route reliably preserves both global semantics and fine-grained evidence in a compact representation.

Recently, \citet{zhu2026mure} introduced MURE that combines multi-resolution sampling with token clustering to produce visual embeddings for effective and efficient document retrieval. 
However, because MURE's clustering is non-trainable and post-hoc, its merge criterion receives no loss feedback, risking the loss of relevant representations and the persistence of irrelevant ones after compression.
Training-time merging and compression-aware fine-tuning have been shown to mitigate such performance degradation~\citep{bolya2023tome,ma2025towards}, suggesting that compression and representations should be jointly optimized rather than treated as separate stages. Motivated by this, we aim to generate a compact, universal visual representation via trainable coarse-to-fine compression that preserves complementary global and local evidence within contextualized representations. Specifically, within each granularity, compression should merge semantically similar tokens while retaining informative content; across granularities, it should reduce redundancy and prioritizes details absent from coarser representations.

In light of this, we propose \textsc{ResComEmb}, a trainable framework for effective and efficient universal multimodal embedding. Specifically, \textsc{ResComEmb} first uses an MLLM (e.g., Qwen2.5-VL's ~\citep{Qwen25-VL}) to encode global, aspect-ratio-aware intermediate, and fine-grained views, providing coarse-to-fine visual evidence. Residual Homogeneity Compression (RHC) then processes the fully contextualized and embedding-projected outputs, consolidating repeated evidence within each granularity while prioritizing new evidence from finer granularities under explicit token budgets. 
Matryoshka Representation Learning (MRL)~\citep{kusupati2022matryoshka} further applies nested supervision to the accumulated coarse-to-fine prefixes, training every compressed representation to remain semantically effective. 
Then, a length-adaptive Bidirectional Late-Interaction Matching is applied for robust query--document scoring, mitigating weak-match accumulation by averaging the strongest token-level matches in both directions and balancing their scores by valid-token counts.

We evaluate \textsc{ResComEmb} on general multimodal tasks using the Massive Multimodal Embedding Benchmark (MMEB) \citep{jiang2025vlmvec}, and on visual document retrieval (VDR) task using ViDoRe V1 \citep{ColPali} and V2 \citep{mace2025vidore}.
Extensive experimental results show that \textsc{ResComEmb} achieves the best average performance across all three benchmarks. It reaches 67.4 Precision@1 on MMEB, exceeding VLM2Vec-V2 by 2.5 points, and 90.4 and 61.6 NDCG@5 on ViDoRe V1 and V2, respectively, surpassing ColQwen2.5 by 1.0 point on each benchmark. Notably, these ViDoRe gains are achieved using only 37.5\% of full-token ColQwen2.5's visual-token budget, demonstrating a favorable effectiveness--efficiency trade-off.
Ablation studies confirm the contributions of each mechanism in \textsc{ResComEmb}, including Residual Homogeneity Compression (RHC), MRL-based nested supervision, and Bidirectional Late-Interaction Matching.

In summary, the key contributions of this work are threefold:
\begingroup
\setlength{\leftmargini}{1.0em}
\begin{itemize}
    \setlength{\itemsep}{1pt}
    \setlength{\parsep}{0pt}
    \setlength{\topsep}{2pt}
    \item  We propose a coarse-to-fine principle for generating universal multimodal embeddings under a certain visual-token budget: gather complementary evidence across multiple granularities while reducing within-granularity redundancy and cross-granularity repetition.
    \item We develop \textsc{ResComEmb}, a trainable framework for effective and efficient multimodal embeddings, which merges redundant tokens within each granularity and prioritizes finer-grained evidence not already captured by coarser representations, subject to a given visual-token budget.
    \item Extensive experiments demonstrate that \textsc{ResComEmb} outperforms VLM2Vec-V2 on MMEB and full-token ColQwen2.5 on ViDoRe V1/V2 while using only 37.5\% of the latter's visual-token budget, establishing a favorable effectiveness--efficiency trade-off.
\end{itemize}
\endgroup

\section{Related Work}
\label{sec:related}

\subsection{Multimodal Embedding}

Multimodal embedding maps images and text into a shared representation space for measuring semantic relevance~\citep{radford2021learning,jiang2025vlmvec}. This field has progressed from contrastive vision--language pre-training to universal MLLM-based representations. CLIP~\citep{radford2021learning} and SigLIP~\citep{zhai2023sigmoid} use modality-specific encoders for global alignment, whereas BLIP~\citep{li2022blip} and CoCa~\citep{yu2022coca} add generative language modeling. However, globally pooled retrieval embeddings do not preserve explicit local visual--text correspondences~\citep{yao2022filip}. MLLMs instead process interleaved image--text sequences in a unified backbone, producing contextualized representations for instruction-conditioned multimodal embedding. E5-V~\citep{jiang2024e5v} uses task-specific prompts, VLM2Vec~\citep{jiang2025vlmvec} and VLM2Vec-V2~\citep{meng2025vlm2vecv2} use instruction-guided contrastive learning, and GME~\citep{zhang2024gme} uses synthesized fused-modal data. Most methods encode each input as a single vector, reducing storage and comparison costs but potentially suppressing localized and compositional evidence~\citep{thrush2022winoground}. Multi-vector methods retain independently accessible features, keeping regions, objects, and text separately encoded rather than prematurely pooled~\citep{khattab2020colbert,yao2022filip}. Recent MLLM-based approaches extend this design to visually rich, interleaved inputs~\citep{ColPali,zhu2026mure}. Building on hierarchical multi-resolution encoding, \textsc{ResComEmb} introduces trainable Residual Homogeneity Compression to construct compact, input-aligned multi-vector embeddings.

\subsection{Visual Representation Compression}
\label{sec:related_compression}

Visual representation compression shortens visual sequences while retaining task-relevant content~\citep{visionzip2024,ma2025towards}. Compression occurs either before or after full MLLM contextualization. Upstream approaches target encoding and generation costs: InternVL 1.5~\citep{chen2024far}, TokenPacker~\citep{li2025tokenpacker}, and LLaVA-UHD v2~\citep{zhang2026llava} construct compact visual inputs through spatial or cross-scale aggregation, whereas VisionZip~\citep{visionzip2024}, SCOPE~\citep{deng2026scope}, and FOLDER~\citep{folder2025} apply content-aware selection or merging. These approaches reduce the number of visual tokens processed within the MLLM. Downstream approaches compress contextualized embeddings for storage and comparison. Token Pooling~\citep{clavie2024reducing} and Light-ColPali~\citep{ma2025towards} merge or cluster visual embeddings, whereas AGC~\citep{qin2026multi} and SaMer~\citep{park2026all} learn representative centers. MetaEmbed~\citep{metaembed} controls length with learnable abstract tokens, enabling global aggregation but potentially omitting input-specific local evidence under tight budgets~\citep{cha2024honeybee,liu2024visual}. MURE~\citep{zhu2026mure} retains input-aligned multi-resolution evidence, but its compression stage relies on non-trainable clustering and does not explicitly distinguish within-granularity redundancy from cross-granularity repetition. By contrast, \textsc{ResComEmb} introduces RHC as a downstream-trainable compressor for ordered multi-granularity MLLM representations, separately reducing these two forms of redundancy under explicit budgets without modifying the backbone.

\section{Method}
\label{sec:method}
\begin{figure}[t]
  \centering
  \vspace{0pt}
  \includegraphics[width=\linewidth]{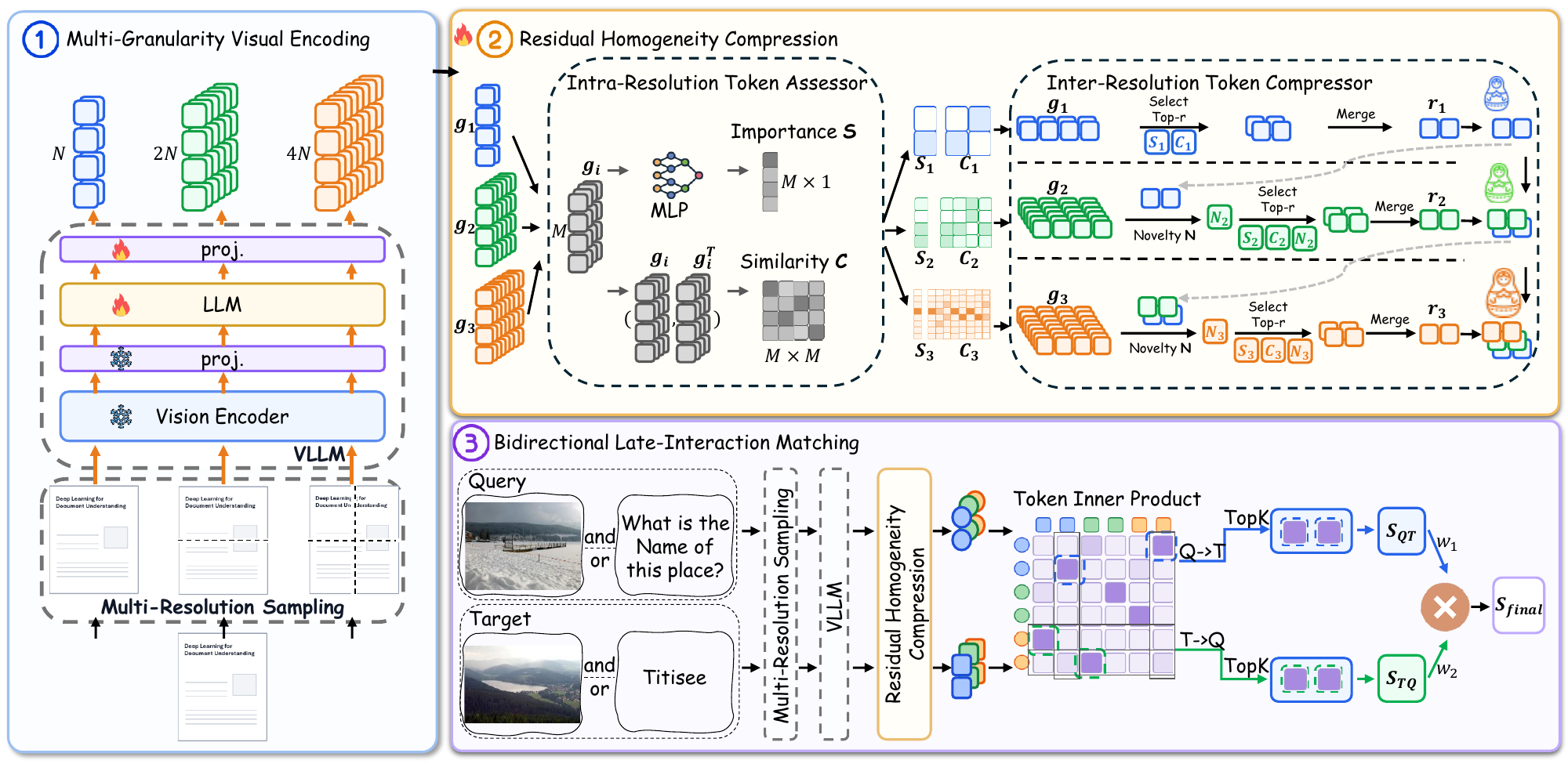}
  \caption{Overview of proposed \textsc{ResComEmb} framework. Ordered multi-granularity views are encoded by a shared MLLM, then compressed into nested representations. The Bidirectional Late-Interaction Matching is applied to compute relevance scores between multimodal queries and targets.}
  \label{fig:comresemb-overview}
\end{figure}

In this section, we first present our problem formulation for budgeted multimodal embedding, and then introduce our \textsc{ResComEmb} framework, as illustrated in Figure~\ref{fig:comresemb-overview}, which comprises the following mechanisms: Multi-Granularity Visual Encoding, Residual Homogeneity Compression, Nested Representation, and Bidirectional Late-Interaction Matching.

\subsection{Problem Formulation}

We formulate budgeted multimodal embedding as learning a mapping from textual and visual inputs, $x^{\mathrm{t}}$ and $x^{\mathrm{v}}$, to a compact sequence of token-level embeddings under a visual-token budget $B_{\mathrm{v}}$:

\begin{equation}
  \mathcal{F}(x^{\mathrm{t}},x^{\mathrm{v}};B_{\mathrm{v}}) \longmapsto
  \mathbf{E}_{(x^{\mathrm{t}},x^{\mathrm{v}})}^{(B_{\mathrm{v}})}\in\mathbb{R}^{(N_{\mathrm{t}}+M_{\mathrm{v}}(B_{\mathrm{v}}))\times d},
  \label{eq:problem-formulation}
\end{equation}
where $N_{\mathrm{t}}$ and $M_{\mathrm{v}}(B_{\mathrm{v}})$ denote the numbers of text and visual tokens, respectively, and $d$ is the embedding dimension. The output $\mathbf{E}_{(x^{\mathrm{t}},x^{\mathrm{v}})}^{(B_{\mathrm{v}})}$ consists of normalized token-level embeddings. All $N_{\mathrm{t}}$ text-token embeddings are retained without budget-based compression, whereas the visual tokens are compressed such that $M_{\mathrm{v}}(B_{\mathrm{v}})\le B_{\mathrm{v}}$.

\subsection{Multi-Granularity Visual Encoding}

For each visual input, Multi-Granularity Visual Encoding constructs views at three granularities: a global $1\times1$ view, an aspect-ratio-aware intermediate view partitioned into $1\times2$ or $2\times1$ regions, and a fine-grained view partitioned into $2\times2$ regions. The three granularities produce $(c_1,c_2,c_3)=(1,2,4)$ crops. The shared MLLM encodes textual inputs together with all visual crops. A learned embedding projection is applied to the last-layer hidden states. The valid projected tokens are separated into text tokens $\mathbf{T}\in\mathbb{R}^{M_t\times d}$ and visual sequences $\mathbf{g}_1,\mathbf{g}_2,\mathbf{g}_3$, where $\mathbf{g}_k\in\mathbb{R}^{M_k\times d}$ and $M_k$ denotes the number of visual tokens at stage $k$. Text-only inputs retain $\mathbf{T}$ and skip visual compression.

\subsection{Residual Homogeneity Compression}

Residual Homogeneity Compression (RHC) takes the three visual token sequences $\mathbf{g}_1,\mathbf{g}_2,\mathbf{g}_3$ as input and processes them under explicit stage budgets. At each stage, an Intra-Resolution Assessor estimates token importance and within-stage similarity. Then Inter-Resolution Compressor combines these signals with outputs from preceding, coarser stages to merge redundant tokens while preserving evidence not yet represented. The compressed sequences are used to construct nested representations.

At stage $k$, RHC processes only the current sequence $\mathbf{g}_k$. The outputs of the preceding stages, $\mathbf{r}_1,\ldots,\mathbf{r}_{k-1}$, are concatenated to form the coarse-anchor set $\mathbf{A}_k$, which is used to estimate cross-stage novelty. The first stage has no preceding anchors.

\paragraph{Intra-Resolution Token Assessor.}
The assessor takes the projected tokens of the visual stage and produces importance and within-stage similarity scores for the subsequent compressor. At stage $k$, a MLP-based contextual encoder is applied to transform the projected visual-token sequence $\mathbf{g}_k$ into contextualized embeddings $\mathbf{h}_k$. The importance score $\mathcal{S}_i$ for each token $i$ is obtained by

\begin{equation}
  \mathcal{S}_i=\operatorname{MinMax}\!\left(
  f_{\mathrm{imp}}(\mathbf{h}_i)\right),
  \label{eq:rhc-importance-score}
\end{equation}
where $f_{\mathrm{imp}}$ is the importance scorer, and $\operatorname{MinMax}$ normalizes scores over tokens in the current stage. To estimate within-stage redundancy, we divide alternating positions into source tokens, $\mathbf{h}_A=\mathbf{h}_{::2}$, and destination tokens, $\mathbf{h}_B=\mathbf{h}_{1::2}$. The similarity score $\mathcal{C}_{ij}$ is their cosine similarity:

\begin{equation}
  \mathcal{C}_{ij}
  =\frac{\mathbf{h}_{A,i}^{\top}\mathbf{h}_{B,j}}
  {\lVert\mathbf{h}_{A,i}\rVert_2\lVert\mathbf{h}_{B,j}\rVert_2}.
  \label{eq:rhc-similarity}
\end{equation}
The assessor then passes $\mathcal{S}$ and $\mathcal{C}$ to the Inter-Resolution Compressor.

\paragraph{Inter-Resolution Token Compressor.}

The Inter-Resolution Compressor combines the assessor outputs with coarse-stage anchors to remove redundant evidence while preserving information introduced at finer stages. For each current-stage token, the novelty score $\mathcal{N}_{i}$ quantifies the evidence not covered by the accumulated anchors:

\begin{equation}
  \mathcal{N}_i=
  \begin{cases}
    1, & k=1,\\[2pt]
    \operatorname{MinMax}\!\left(
    1-\displaystyle\max_{\mathbf{a}\in\mathbf{A}_k}
    \frac{\mathbf{h}_i^\top\mathbf{a}}
    {\lVert\mathbf{h}_i\rVert_2\lVert\mathbf{a}\rVert_2}
    \right), & k>1.
  \end{cases}
  \label{eq:rhc-residual-gain}
\end{equation}

The novelty score $\mathcal{N}$ and importance score $\mathcal{S}$ jointly define the preservation priority vector, $\mathcal{P}_{i}=\mathcal{S}_{i}+\eta\mathcal{N}_{i}$, where $\eta$ controls the contribution of cross-stage novelty. The compressor assigns each candidate source--destination pair a merge score $\mathcal{U}_{ij}$ based on its similarity $\mathcal{C}_{ij}$ and the source token's preservation priority $\mathcal{P}_i$:
\[
  \mathcal{U}_{ij}=\mathcal{C}_{ij}-\alpha\mathcal{P}_{i},
\]
where $\alpha$ controls the extent to which preservation priority offsets similarity. 
Specifically, each source token is assigned to its highest-scoring destination, and merges are performed until the output meets the stage-specific token budget. The token values assigned to each destination are then aggregated and $\ell_2$-normalized to obtain $\mathbf{r}_k$. Although the score assignments are discrete, value aggregation remains differentiable. Preservation priorities govern source selection rather than destination assignment: sources are ranked by their best merge scores, making high-priority tokens less likely to be selected for merging. These priority signals also modulate value aggregation, enabling gradients to train the assessor.The detailed derivation is provided in Appendix~\ref{sec:appendix-rhc-gradient}.

\paragraph{Nested Representation.}
Then, Nested Representation assembles the surviving outputs $\mathbf{r}_1,\ldots,\mathbf{r}_n$ in coarse-to-fine order. In accordance with the Matryoshka principle~\citep{kusupati2022matryoshka}, the assembled sequence supports multiple representation budgets through its prefixes. For an input $x$, the level-$k$ representation is defined as
\begin{equation}
  \mathbf{E}_{x,k}
  =\operatorname{Concat}\!\left(
  \mathbf{T}_x,\mathbf{r}_{x,1},\ldots,
  \mathbf{r}_{x,k}\right),
  \qquad k\in\{1,\ldots,n\},
  \label{eq:granularity-nesting}
\end{equation}
For each $k<n$, $\mathbf{E}_{x,k}$ is a prefix of $\mathbf{E}_{x,k+1}$. All levels retain the same text prefix and append progressively finer compressed visual stages. The nested representations are generated for both queries and documents using the same procedure.

\subsection{Bidirectional Late-Interaction Matching}

Given the nested representations, Bidirectional Late-Interaction Matching computes a query--document score using  similarities in both directions. Let $\mathbf{E}_q=[\mathbf{e}_q^i]_{i=1}^{N_q}$ and $\mathbf{E}_d=[\mathbf{e}_d^j]_{j=1}^{N_d}$ denote the query and document token sequences, with $N_q$ and $N_d$ denoting respective numbers of valid tokens.

\begin{equation}
  K_q=\min(K,N_q),\qquad K_d=\min(K,N_d).
  \label{eq:topk-counts}
\end{equation}
Let $\Omega_q$ contain the indices of the $K_q$ query tokens with the highest maximum similarities to any document token. Analogously, let $\Omega_d$ contain the indices of the $K_d$ document tokens with the highest maximum similarities to any query token. The bidirectional Top-$K$ means are defined as

\begin{equation}
\begin{aligned}
  S_{QT}^{(K)}
  &=\frac{1}{K_q}\sum_{i\in\Omega_q}
    \max_{1\le j\le N_d}(\mathbf{e}_q^i)^\top\mathbf{e}_d^j,
  \qquad
  S_{TQ}^{(K)}
  &=\frac{1}{K_d}\sum_{j\in\Omega_d}
    \max_{1\le i\le N_q}(\mathbf{e}_q^i)^\top\mathbf{e}_d^j,
\end{aligned}
\label{eq:bidirectional-topk-mean}
\end{equation}

The final query--document score combines the two directional scores as

\begin{equation}
  S_{\mathrm{final}}(\mathbf{E}_q,\mathbf{E}_d)
  =w_1(q,d)S_{QT}^{(K)}+w_2(q,d)S_{TQ}^{(K)},
\label{eq:adaptive-score}
\end{equation}
where $S_{QT}^{(K)}$ measures query-to-document support and $S_{TQ}^{(K)}$ measures document-to-query coverage. The adaptive weights $w_1(q,d)$ and $w_2(q,d)$ balance the two directions according to token counts. When the query and document have equal valid-token counts, both directions receive equal weight. As the document becomes longer, the weighting shifts toward query-to-document support while retaining a contribution from reverse coverage. Appendix~\ref{sec:appendix-interaction} provides more detailed implementation.

\subsection{Training}

The matching score provides the training signal for each active nested representation. Consider a batch $\mathcal{B}=\{(q_i,d_i,\mathcal{D}_i^-)\}_{i=1}^{|\mathcal{B}|}$, where $\mathcal{D}_i^-$ is an optional set of explicit negatives. For level $k$, let $\mathcal{I}_k$ index the examples for which that level is active. For each $i\in\mathcal{I}_k$, define the candidate set as $\mathcal{C}_{i,k}=\{d_j:j\in\mathcal{I}_k\}\cup\mathcal{D}_i^-$. The score of any candidate $d_j\in\mathcal{C}_{i,k}$ is $s_{ij}^{(k)}=S_{\mathrm{final}}(\mathbf{E}_{q_i,k},\mathbf{E}_{d_j,k})$.
Using these candidate scores, we define the level-wise contrastive loss and the overall training objective as follows:
\begin{equation}
\begin{aligned}
  \mathcal{L}_{k}
  &=-\frac{1}{|\mathcal{I}_k|}
  \sum_{i\in\mathcal{I}_k}
  \log
  \frac{\exp\!\left(s_{ii}^{(k)}/\tau\right)}
  {\displaystyle\sum_{d_j\in\mathcal{C}_{i,k}}
   \exp\!\left(s_{ij}^{(k)}/\tau\right)}
  ,\qquad
  \mathcal{L}_{\mathrm{train}}
  =\sum_{k\in\mathcal{K}_{\mathcal{B}}}\lambda_k\mathcal{L}_k,
\end{aligned}
\label{eq:training-objective}
\end{equation}
where $\mathcal{K}_{\mathcal{B}}$ is the set of levels active for at least one example in the batch, $\tau$ is the contrastive temperature, and $\lambda_k$ weights the loss at level $k$. Each $\mathcal{C}_{i,k}$ contains the in-batch documents associated with active examples and any available explicit negatives.

\section{Experiments}
\label{sec:experiments}

We first present the experimental setup and main results on MMEB and ViDoRe V1/V2, followed by ablation studies and analysis.

\subsection{Experimental Setup}

\paragraph{Evaluation Datasets.}
We evaluate our \textsc{ResComEmb} in general multimodal embedding tasks on MMEB~\citep{jiang2025vlmvec} and the visual document retrieval task on ViDoRe V1~\citep{ColPali} and ViDoRe V2~\citep{mace2025vidore}, reporting Precision@1 for MMEB and NDCG@5 for both ViDoRe benchmarks. More details are provided in Appendix~\ref{sec:appendix-setup}.

\paragraph{Compared Methods.}
We compare single- and multi-vector models. Single-vector baselines include CLIP~\citep{radford2021learning}, SigLIP~\citep{zhai2023sigmoid}, ONE-PEACE~\citep{wang2023onepeace}, UniIR~\citep{wei2024uniir}, MagicLens~\citep{zhang2024magiclens}, E5-V~\citep{jiang2024e5v}, VLM2Vec~\citep{jiang2025vlmvec}, VLM2Vec-V2~\citep{meng2025vlm2vecv2}, GME~\citep{zhang2024gme}, LamRA~\citep{liu2025lamra}, MMRet~\citep{zhou2025megapairs}, and mmE5~\citep{chen2025mme5}; DSE~\citep{ma2024dse} and VisRAG-Ret~\citep{visrag} provide document-focused comparisons. Multi-vector baselines are ColPali and ColQwen2~\citep{ColPali}, which extend late interaction to visual document patches; ColQwen2.5, which applies it to Qwen2.5-VL~\citep{ColPali,Qwen25-VL}; ColMate-Pali and ColMate~\citep{masry-etal-2025-colmate}, which add contrastive late interaction and masked-text supervision; MetaEmbed~\citep{metaembed}, which uses learnable tokens for compact vector sets; and MURE~\citep{zhu2026mure}, which combines multi-resolution encoding with token clustering.
\subsection{Main Results}

We compare the overall performance of our proposed \textsc{ResComEmb} method with baseline methods on the multimodal embedding benchmark MMEB in Table~\ref{tab:mmeb}, and on the visual document retrieval benchmarks ViDoRe V1 and V2 in Tables~\ref{tab:qwen25-vidore-v1} and~\ref{tab:vidore}, respectively. 
All \textsc{ResComEmb} entries use per-stage RHC budgets of $128/128/128$(384 in total). 
We summarize the main observations below.

\begin{table}[t]
  \centering
  \caption{MMEB Precision@1 (\%). Cls., Ret., and Gnd.\ abbreviate classification, retrieval, and grounding; IND/OOD denote in-domain/out-of-domain tasks. Bold and underline mark the best and second-best scores in each column.}
  \label{tab:mmeb}
  \vspace{3pt}
  \small
  \setlength{\tabcolsep}{1pt}
  \renewcommand{\arraystretch}{1.08}
  \begin{tabular*}{\linewidth}{@{}l@{\hspace{4pt}}l@{\hspace{4pt}\extracolsep{\fill}}lrrrrrrr@{}}
    \toprule
    \multirow{2}{*}{\textbf{Model}} & \multirow{2}{*}{\textbf{Backbone}} & \multirow{2}{*}{\textbf{Size}} & \multicolumn{4}{c}{\textbf{Per Meta-Task Score}} & \multicolumn{3}{c}{\textbf{Average Score}} \\
    \cmidrule(lr){4-7} \cmidrule(l){8-10}
    & & & \textbf{Cls.} & \textbf{VQA} & \textbf{Ret.} & \textbf{Gnd.} & \textbf{IND} & \textbf{OOD} & \textbf{Avg.} \\
    \midrule
    \multicolumn{10}{c}{\textbf{\emph{Single-Vector Embedding}}} \\
    CLIP & ViT-L & 428M & 55.2 & 19.7 & 53.2 & 62.2 & 47.6 & 42.8 & 45.4 \\
    UniIR & ViT-L & 428M & 44.3 & 16.2 & 61.8 & 65.3 & 47.1 & 41.7 & 44.7 \\
    MagicLens & ViT-L & 613M & 38.8 & 8.3 & 35.4 & 26.0 & \multicolumn{1}{c}{--} & \multicolumn{1}{c}{--} & 27.8 \\
    VLM2Vec & Qwen2-VL & 2B & 58.7 & 49.3 & 65.0 & 72.9 & 64.9 & 53.3 & 59.7 \\
    VLM2Vec-V2 & Qwen2-VL & 2B & \underline{62.9} & 56.3 & 69.5 & 77.3 & 68.8 & \underline{59.9} & 64.9 \\
    GME & Qwen2-VL & 2B & 54.4 & 29.9 & 66.9 & 55.5 & 49.2 & 55.2 & 51.9 \\
    GME & Qwen2-VL & 7B & 57.7 & 34.7 & \textbf{71.2} & 59.3 & 53.6 & 58.9 & 56.0 \\
    VLM2Vec & Qwen2-VL & 7B & 62.7 & 56.9 & 69.4 & 82.2 & \underline{71.4} & 58.1 & \underline{65.5} \\
    LamRA & Qwen2-VL & 7B & 59.2 & 26.5 & \underline{70.0} & 62.7 & 53.0 & 55.4 & 54.1 \\
    LamRA & Qwen2.5-VL & 7B & 51.7 & 34.1 & 66.9 & 56.7 & 51.7 & 53.3 & 52.4 \\
    MMRet & LLaVA-1.6 Mistral & 7B & 56.0 & \underline{57.4} & 69.9 & \underline{83.6} & 68.0 & 59.1 & 64.1 \\
    \midrule
    \multicolumn{10}{c}{\textbf{\emph{Multi-Vector Embedding}}} \\
    ColPali & PaliGemma & 3B & 40.3 & 11.5 & 48.1 & 40.3 & 35.0 & 34.7 & 34.9 \\
    \midrule
    \textbf{\textsc{ResComEmb}} & Qwen2.5-VL & 3B & \textbf{63.1} & \textbf{61.1} & 69.5 & \textbf{87.9} & \textbf{71.7} & \textbf{62.1} & \textbf{67.4} \\
    \bottomrule
  \end{tabular*}
\end{table}

\paragraph{Overall Performance Comparison on MMEB.}
\textsc{ResComEmb} achieves the highest MMEB average of 67.4. At a comparable model scale, it outperforms VLM2Vec-V2 by 2.5 points, while also exceeding the larger VLM2Vec-7B by 1.9 points. \textsc{ResComEmb} further leads both the in-domain and out-of-domain averages, showing consistent performance across evaluation domains.

\begin{table}[t]
  \centering
  \caption{ViDoRe V1 NDCG@5 (\%) across 10 visual document retrieval tasks. Arxiv, Doc, and Info denote ArxivQ, DocQ, and InfoQ; Ener.\ and Hlth.\ abbreviate Energy and Health.}
  \label{tab:qwen25-vidore-v1}
  \vspace{3pt}
  \small
  \setlength{\tabcolsep}{0.8pt}
  \renewcommand{\arraystretch}{1.08}
  \begin{tabular*}{\linewidth}{@{}l@{\hspace{2pt}}l@{}c@{\hspace{2pt}\extracolsep{\fill}}*{11}{R{19.5pt}}@{}}
    \toprule
    \textbf{Model} & \textbf{Backbone} & \textbf{Size} & \makebox[19.5pt][c]{\hspace{2pt}\textbf{Arxiv}} & \makebox[19.5pt][c]{\hspace{2pt}\textbf{Doc}} & \makebox[19.5pt][c]{\hspace{2pt}\textbf{Info}} & \makebox[19.5pt][c]{\hspace{2pt}\textbf{TabF}} & \makebox[19.5pt][c]{\hspace{2pt}\textbf{TATQ}} & \makebox[19.5pt][c]{\hspace{4pt}\textbf{Shift}} & \makebox[19.5pt][c]{\textbf{AI}} & \makebox[19.5pt][c]{\textbf{Ener.}} & \makebox[19.5pt][c]{\textbf{Gov.}} & \makebox[19.5pt][c]{\textbf{Hlth.}} & \makebox[19.5pt][c]{\textbf{Avg.}} \\
    \midrule
    \multicolumn{14}{c}{\textbf{\emph{Single-Vector Embedding}}} \\
    ONE-PEACE & ONE-PEACE & 4B & 43.9 & 23.4 & 59.9 & 57.0 & 13.4 & 17.0 & 45.4 & 53.2 & 55.9 & 59.5 & 42.9 \\
    E5-V & LLaVA-NeXT & 8B & 41.1 & 24.3 & 49.5 & 58.2 & 9.0 & 13.2 & 46.1 & 57.7 & 53.0 & 59.6 & 41.2 \\
    DSE & Phi-3-Vision & 4B & 78.1 & 45.8 & 82.0 & 79.2 & 49.0 & 69.8 & 96.8 & 92.6 & 92.0 & 96.3 & 78.2 \\
    GME & Qwen2-VL & 2B & 82.8 & 53.1 & 90.2 & 93.3 & 69.9 & 89.5 & 97.5 & 91.9 & 94.6 & 98.7 & 86.2 \\
    GME & Qwen2-VL & 7B & 86.9 & 57.5 & 91.6 & \underline{94.6} & 74.1 & \textbf{96.8} & \textbf{99.6} & 95.3 & \textbf{98.8} & \textbf{99.3} & 89.5 \\
    LamRA & Qwen2.5-VL & 7B & 53.0 & 25.4 & 72.3 & 66.1 & 25.9 & 27.3 & 72.0 & 65.2 & 72.2 & 83.8 & 56.3 \\
    VLM2Vec & Qwen2-VL & 2B & 48.9 & 27.0 & 67.2 & 62.6 & 19.8 & 41.8 & 55.0 & 59.1 & 57.1 & 59.6 & 49.8 \\
    VLM2Vec & Qwen2-VL & 7B & 60.2 & 34.7 & 70.4 & 78.2 & 27.6 & 38.6 & 67.7 & 60.4 & 61.8 & 69.9 & 57.0 \\
    VLM2Vec-V2 & Qwen2-VL & 2B & 80.6 & 44.9 & 83.7 & 89.2 & 43.8 & 60.8 & 88.5 & 86.5 & 85.0 & 92.2 & 75.5 \\
    \midrule
    \multicolumn{14}{c}{\textbf{\emph{Multi-Vector Embedding}}} \\
    ColPali v1.3 & PaliGemma & 3B & 81.7 & 56.6 & 84.9 & 86.9 & 70.9 & 75.1 & 95.7 & 94.7 & 93.6 & 95.9 & 83.6 \\
    ColMate-Pali & PaliGemma & 3B & 83.6 & 57.5 & 84.1 & 87.6 & 74.0 & 79.8 & 98.3 & 94.1 & 95.3 & 96.6 & 85.1 \\
    MURE & PaliGemma & 3B & 84.6 & \underline{61.7} & 89.0 & 89.3 & 76.8 & 83.2 & 98.7 & 95.2 & 94.4 & 97.1 & 87.0 \\
    ColQwen2 & Qwen2-VL & 2B & 88.0 & 61.5 & 92.5 & 89.0 & \textbf{82.2} & 89.9 & 99.0 & 95.9 & 95.5 & \underline{98.8} & 89.2 \\
    ColQwen2.5 & Qwen2.5-VL & 3B & \underline{89.2} & \textbf{63.2} & 92.4 & 91.1 & 81.1 & 87.3 & \textbf{99.6} & 95.9 & 96.4 & 97.9 & 89.4 \\
    ColMate & Qwen2.5-VL & 3B & \textbf{90.2} & 61.1 & \underline{93.7} & 91.5 & \underline{81.9} & 90.2 & \underline{99.3} & \underline{96.4} & 96.5 & 98.1 & \underline{89.9} \\
    \midrule
    \textbf{\textsc{ResComEmb}} & Qwen2.5-VL & 3B & 88.8 & \underline{61.7} & \textbf{94.2} & \textbf{95.2} & 80.4 & \underline{90.4} & \textbf{99.6} & \textbf{96.6} & \underline{97.9} & \textbf{99.3} & \textbf{90.4} \\
    \bottomrule
  \end{tabular*}
\end{table}

\begin{table}[t]
  \centering
  \caption{ViDoRe V2 NDCG@5 (\%) across 7 tasks. Syn, Mul, and Bio denote synthetic, multilingual, and biomedical data.}
  \label{tab:vidore}
  \vspace{3pt}
  \small
  \setlength{\tabcolsep}{1.5pt}
  \renewcommand{\arraystretch}{1.08}
  \begin{tabular*}{\linewidth}{@{}l@{\hspace{2pt}}l@{\hspace{1pt}}c@{\hspace{3pt}\extracolsep{\fill}}*{8}{R{18pt}}@{}}
    \toprule
    \textbf{Model} & \textbf{Backbone} & \textbf{Size} & \makebox[18pt][c]{\shortstack{\textbf{ESG}\\\textbf{Human}}} & \makebox[18pt][c]{\shortstack{\textbf{Eco}\\\textbf{Mul}}} & \makebox[18pt][c]{\shortstack{\textbf{Bio}\\\textbf{Mul}}} & \makebox[18pt][c]{\shortstack{\textbf{ESG}\\\textbf{Syn-Mul}}} & \makebox[18pt][c]{\textbf{Bio}} & \makebox[18pt][c]{\shortstack{\textbf{ESG}\\\textbf{Syn}}} & \makebox[18pt][c]{\textbf{Eco}} & \makebox[18pt][c]{\textbf{Avg.}} \\
    \midrule
    \multicolumn{11}{c}{\textbf{\emph{Single-Vector Embedding}}} \\
    SigLIP & SigLIP & 652M & 28.8 & 14.0 & 18.2 & 21.9 & 33.8 & 19.8 & 29.8 & 23.8 \\
    \addlinespace[2pt]
    VisRAG-Ret & MiniCPM-V2.0 & 3B & 53.7 & 48.7 & 47.7 & 46.4 & 54.8 & 45.9 & 59.6 & 51.0 \\
    VLM2Vec & Qwen2-VL & 7B & 33.9 & 42.0 & 29.7 & 38.4 & 38.8 & 36.7 & 51.4 & 38.7 \\
    GME & Qwen2-VL & 7B & 65.8 & \underline{56.2} & 55.1 & 56.7 & \underline{64.0} & 54.3 & \textbf{62.9} & 59.3 \\
    \addlinespace[2pt]
    mmE5 & Llama-3.2-Vision & 11B & 52.8 & 44.3 & 46.8 & 54.7 & 51.3 & 55.1 & 48.6 & 50.5 \\
    \midrule
    \multicolumn{11}{c}{\textbf{\emph{Multi-Vector Embedding}}} \\
    ColPali & PaliGemma & 3B & 51.1 & 49.9 & 56.5 & 55.7 & 59.7 & 57.0 & 51.6 & 54.5 \\
    ColMate-Pali & PaliGemma & 3B & 62.8 & 54.1 & 59.3 & 53.4 & 60.9 & 54.1 & 55.9 & 57.2 \\
    MURE & PaliGemma & 3B & 67.9 & 54.5 & 56.6 & \underline{57.4} & 60.4 & \underline{62.4} & 57.3 & 59.5 \\
    \addlinespace[2pt]
    ColQwen2 & Qwen2-VL & 2B & 62.2 & 53.2 & 56.5 & 54.2 & 61.8 & 53.4 & 61.5 & 57.5 \\
    MetaEmbed & Qwen2.5-VL & 3B & 63.7 & 55.5 & 58.7 & \underline{57.4} & 61.7 & \textbf{62.6} & 62.3 & 60.3 \\
    ColQwen2.5 & Qwen2.5-VL & 3B & 68.4 & \textbf{56.5} & \textbf{61.1} & \underline{57.4} & 63.6 & 57.4 & 59.8 & \underline{60.6} \\
    ColMate & Qwen2.5-VL & 3B & \underline{68.9} & 52.1 & \underline{60.3} & \textbf{60.2} & 62.1 & 60.1 & 59.6 & 60.5 \\
    \midrule
    \textbf{\textsc{ResComEmb}} & Qwen2.5-VL & 3B & \textbf{69.8} & 56.0 & 60.0 & 57.1 & \textbf{64.8} & 60.9 & \underline{62.7} & \textbf{61.6} \\
    \bottomrule
  \end{tabular*}
\end{table}

\paragraph{Overall Performance Comparison on ViDoRe V1/V2.}
\textsc{ResComEmb} achieves the highest average scores of 90.4 and 61.6 on ViDoRe V1 and V2. Against the strongest specialized baseline methods on each benchmark, it improves over ColMate by 0.5 points on V1 and ColQwen2.5 by 1.0 point on V2, demonstrating consistent gains across both benchmark versions.

\paragraph{\textsc{ResComEmb} Remains Effective under Compact Token Budgets.}
Using the same Qwen2.5-VL-3B backbone, \textsc{ResComEmb} uses 384 visual tokens, 37.5\% of ColQwen2.5 token budget, yet improves average NDCG@5 by 1.0 point on both ViDoRe benchmarks. This demonstrates that our method substantially reduces token budget while maintaining superior retrieval effectiveness.

\subsection{Ablation Studies and In-Depth Analysis}
\label{sec:main-ablations}

We first ablate MRL-based nested supervision and RHC scoring signals, then analyze nested-prefix scaling and RHC token budgets.
Additional analyses of late-interaction scoring, the multi-resolution oracle, and visual-granularity composition appear in Appendices~\ref{sec:appendix-interaction}--\ref{sec:appendix-multigranularity}.

\begin{figure}[!htbp]
  \centering
  \begin{minipage}[t]{0.358\linewidth}
    \centering
    \includegraphics[width=\linewidth]{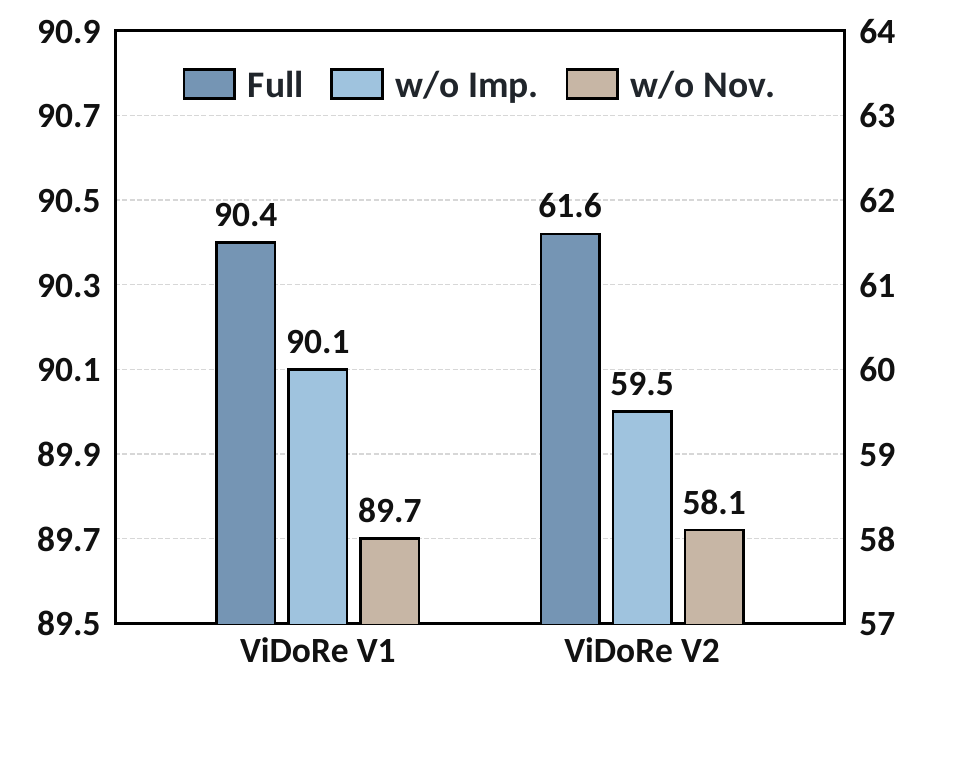}
    {\small\textbf{(a)} RHC scoring signals\par}
  \end{minipage}\hspace{0.006\linewidth}%
  \begin{minipage}[t]{0.315\linewidth}
    \centering
    \includegraphics[width=\linewidth]{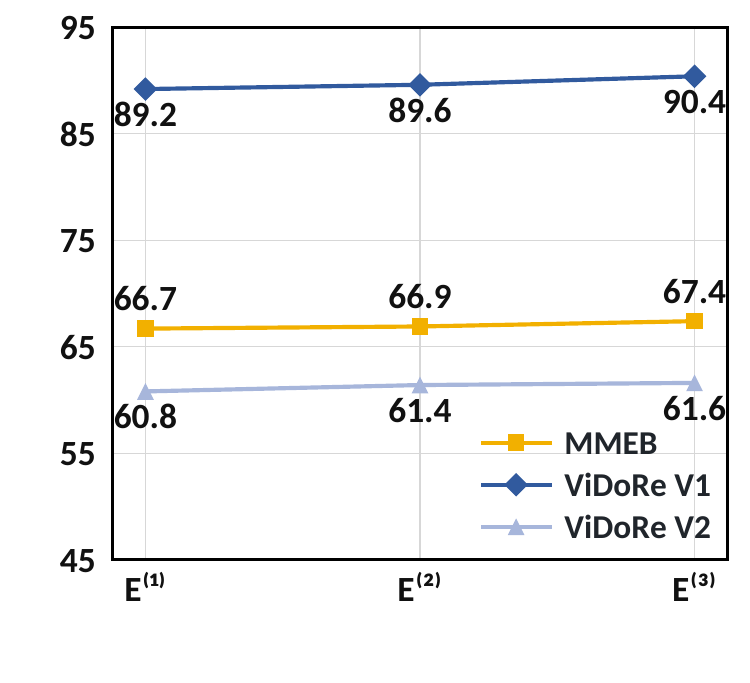}
    {\small\textbf{(b)} Nested supervision\par}
  \end{minipage}\hspace{0.006\linewidth}%
  \begin{minipage}[t]{0.315\linewidth}
    \centering
    \includegraphics[width=\linewidth]{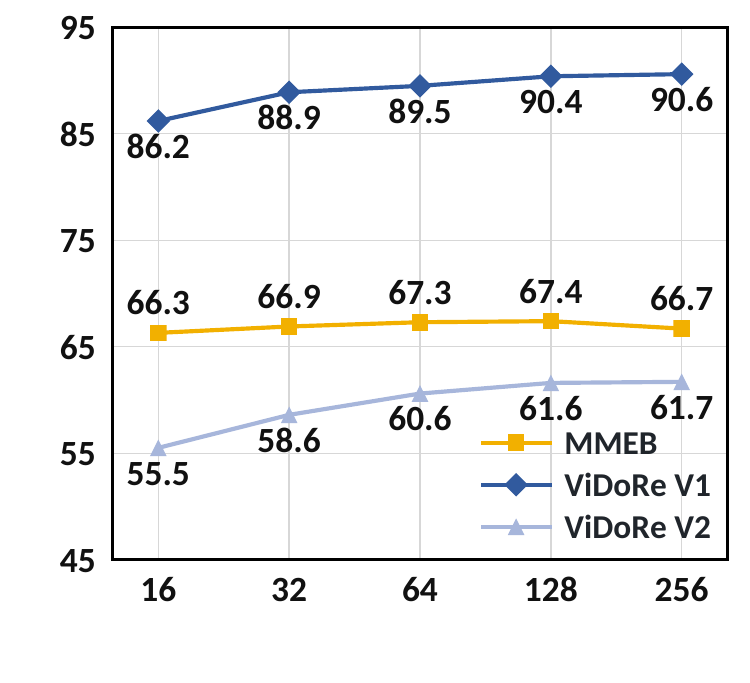}
    {\small\textbf{(c)} RHC token budget\par}
  \end{minipage}
  \caption{Effects of RHC scoring signals, nested-prefix scaling, and RHC token budgets. (a) Full method versus variants without the importance score (w/o Imp.) or the novelty score (w/o Nov.). (b) Three prefixes from one MRL-trained model. (c) Separately trained models across per-stage budgets.}
  \label{fig:main-ablation-plots}
\end{figure}

\paragraph{Ablation on MRL-Based Nested Supervision.}
To quantify the contribution of nested supervision, Table~\ref{tab:mrl-prefix-comparison} compares matched models trained with and without MRL. We can observe that: 1) Removing MRL lowers ViDoRe V1 by 1.0--2.8 points, ViDoRe V2 by 5.7--7.6 points, and MMEB by 2.8--3.0 points across prefixes containing 128, 256, and 384 visual tokens, showing consistent performance gains with MRL at every representation budget. 
2) Across all three prefixes, removing MRL degrades performance on every MMEB task category: classification decreases by 3.6–3.9 points, VQA by 0.5–1.2 points, retrieval by 4.1–5.0 points, and grounding by 1.5–2.5 points. This consistent degradation across heterogeneous task types indicates that nested supervision is essential for preserving task-relevant evidence at each prefix length, rather than benefiting only a subset of tasks.

\paragraph{Ablation on RHC Scoring Signals.}
To isolate each RHC signal's contribution, we retrain matched variants with either the importance or novelty score removed. The experiments results on ViDoRe V1/V2 are presented in Figure~\ref{fig:main-ablation-plots}(a). 
We can observe: 1) Either removal hurts the model performance. Removing the importance score lowers ViDoRe V1 by 0.3 points and V2 by 2.1 points,  while removing the novelty score lowers V1 by 0.7 points and V2 by 3.5 points, confirming their complementary roles in token preservation. 
2) Removing the novelty score causes larger drops on both benchmarks, suggesting its greater contribution to RHC. This is consistent with its role in distinguishing residual fine-grained evidence from content already represented by coarser views, reducing cross-granularity repetition while preserving complementary details in the multi-resolution representation.
More detailed analysis are provided in Appendix Tables~\ref{tab:rhc-component-ablation}--\ref{tab:rhc-component-ablation-v2}.

\begin{table}[!t]
  \centering
  \caption{Effect of MRL across nested prefixes. Prefixes $\mathbf{E}^{(1)}$, $\mathbf{E}^{(2)}$, and $\mathbf{E}^{(3)}$ retain 128, 256, and 384 visual tokens, respectively, with identical text tokens; $\Delta$ denotes the drop without MRL.}
  \label{tab:mrl-prefix-comparison}
  \vspace{3pt}
  \small
  \providecommand{\mrlDrop}[1]{{\scriptsize\textcolor{red}{(#1$\downarrow$)}}}
  \setlength{\tabcolsep}{3pt}
  \renewcommand{\arraystretch}{1.1}
  \begin{tabular*}{\linewidth}{@{\extracolsep{\fill}}lcrrrrrrrr@{}}
    \toprule
    \multirow{2}{*}{\textbf{Prefix}} & \multirow{2}{*}{\textbf{Variant}} & \multicolumn{3}{c}{\textbf{ViDoRe}} & \multicolumn{5}{c}{\textbf{MMEB}} \\
    \cmidrule(lr){3-5}\cmidrule(lr){6-10}
    & & \textbf{V1} & \textbf{V2} & \textbf{Avg.} & \textbf{Cls.} & \textbf{VQA} & \textbf{Ret.} & \textbf{Gnd.} & \textbf{Avg.} \\
    \midrule
    \multirow{3}{*}{$\mathbf{E}^{(1)}$} & w/ MRL & 89.2 & 60.8 & 77.5 & 62.8 & 61.2 & 67.7 & 86.8 & 66.7 \\
    & w/o MRL & 86.4 & 53.2 & 72.7 & 58.9 & 60.0 & 63.6 & 85.1 & 63.7 \\
    & $\Delta$ & \mrlDrop{2.8} & \mrlDrop{7.6} & \mrlDrop{4.8} & \mrlDrop{3.9} & \mrlDrop{1.2} & \mrlDrop{4.1} & \mrlDrop{1.7} & \mrlDrop{3.0} \\
    \midrule
    \multirow{3}{*}{$\mathbf{E}^{(2)}$} & w/ MRL & 89.6 & 61.4 & 78.0 & 62.8 & \textbf{61.4} & 68.2 & 86.8 & 66.9 \\
    & w/o MRL & 88.6 & 55.2 & 74.8 & 59.2 & 60.4 & 64.1 & 85.3 & 64.1 \\
    & $\Delta$ & \mrlDrop{1.0} & \mrlDrop{6.2} & \mrlDrop{3.2} & \mrlDrop{3.6} & \mrlDrop{1.0} & \mrlDrop{4.1} & \mrlDrop{1.5} & \mrlDrop{2.8} \\
    \midrule
    \multirow{3}{*}{$\mathbf{E}^{(3)}$} & w/ MRL & \textbf{90.4} & \textbf{61.6} & \textbf{78.5} & \textbf{63.1} & 61.1 & \textbf{69.5} & \textbf{87.9} & \textbf{67.4} \\
    & w/o MRL & 89.1 & 55.9 & 75.4 & 59.5 & 60.6 & 64.5 & 85.4 & 64.4 \\
    & $\Delta$ & \mrlDrop{1.3} & \mrlDrop{5.7} & \mrlDrop{3.1} & \mrlDrop{3.6} & \mrlDrop{0.5} & \mrlDrop{5.0} & \mrlDrop{2.5} & \mrlDrop{3.0} \\
    \bottomrule
  \end{tabular*}
\end{table}

\paragraph{Performance Analysis of Nested Prefixes.}
To assess elastic inference capability of our \textsc{ResComEmb}, Figure~\ref{fig:main-ablation-plots}(b) evaluates three prefixes from the same MRL-trained model: $\mathbf{E}^{(1)}$, $\mathbf{E}^{(2)}$, and $\mathbf{E}^{(3)}$, which retain 128, 256, and 384 visual vectors, respectively. We can observe: 
1) The 128-token prefix retains 98.7\% of the full 384-token prefix's performance on ViDoRe V1, 98.7\% on ViDoRe V2, and 99.0\% on MMEB, showing that the coarsest prefix captures most task-relevant multimodal evidence and enables low-storage inference without retraining. 
2) From 128 to 384 tokens, performance increases monotonically by 1.2 points on V1, 0.8 points on V2, and 0.7 points on MMEB, indicating that later prefixes progressively add complementary fine-grained evidence and provide a controllable effectiveness--efficiency trade-off of our model.

\paragraph{Performance Analysis of RHC Token Budgets.}
To study the effectiveness--efficiency trade-off, Figure~\ref{fig:main-ablation-plots}(c) compares separately trained models with per-stage budgets of 16, 32, 64, 128, and 256 visual tokens, corresponding to total budgets of 48, 96, 192, 384, and 768 visual tokens. We observe: 
1) Scaling from 48 to 192 vectors improves \textsc{ResComEmb} by 3.3 points on ViDoRe V1, 5.1 points on ViDoRe V2, and 1.0 point on MMEB, showing that capacity matters under tight compression. 
2) Increasing from 192 to 384 vectors yields further gains of 0.9 points on V1, 1.0 point on V2, and 0.1 point on MMEB. Doubling further to 768 vectors adds only 0.2 points on V1 and 0.1 point on V2, while reducing MMEB by 0.7 points. This pattern of diminishing and eventually negative returns indicates that 384 vectors offer the best effectiveness–efficiency balance, motivating the 128/128/128 operating point in this work.
More detailed experimental results are provided in Appendix Table~\ref{tab:appendix-token-budget-ablation}. 

\section{Conclusion}
\label{sec:conclusion}

In this work, we introduce \textsc{ResComEmb} that combines residual compression, nested supervision, and bidirectional late interaction to build compact universal multimodal embeddings. Its trainable coarse-to-fine design reduces within-granularity redundancy and cross-granularity repetition while preserving residual fine-grained evidence under explicit token budgets. It establishes new state-of-the-art results on MMEB and ViDoRe V1/V2, outperforming both universal multimodal retrievers and specialized visual document retrieval models. With only 37.5\% of the visual-token budget of full-token ColQwen2.5, \textsc{ResComEmb} also achieves higher retrieval accuracy on both ViDoRe benchmarks. 
Our results show that disentangling within-granularity redundancy from cross-granularity repetition allows representation size to be substantially reduced while preserving complementary visual information. This perspective suggests that redundancy in multi-vector embeddings is structured, motivating future work on task-adaptive redundancy modeling and the broader application of trainable compression to other representation generation settings.

\subsection*{AI use statement}

In this work, we used generative AI tools to assist with translation and interpretation of experimental results, literature search, manuscript organization, language editing, LaTeX formatting, and drafting the gradient analysis in Appendix~\ref{sec:appendix-rhc-gradient}. We independently checked that analysis against the implementation and verified its gradients with automatic differentiation. We have not used generative AI tools to generate synthetic data sets or experimental results. We have reviewed all AI-assisted work and take responsibility for the final content of this work, including text, claims, or artifacts produced with the aid of generative AI.

\subsection*{Ethics statement}

This work uses publicly available benchmark datasets, including MMEB, ViDoRe V1, and ViDoRe V2, for multimodal representation learning and retrieval evaluation. We do not collect new data from human participants, conduct human-subject experiments, or release personal or sensitive information. The experiments follow the cited benchmark protocols and use these resources for research evaluation. Benchmark composition may reflect domain, language, synthetic-data, or annotation biases; therefore, the reported results should not be interpreted as evidence of fairness or suitability for high-stakes deployment. We are not aware of additional privacy, security, legal, or conflict-of-interest concerns introduced by the method as evaluated in this work.

\subsection*{Reproducibility Statement}

Reproducing our experiments requires only publicly released data resources. The exact Hugging Face repositories used to assemble the training pairs, obtain explicit hard negatives, and conduct evaluation are listed below:
\begingroup
\setlength{\leftmargini}{1.8em}
\begin{itemize}
    \setlength{\itemsep}{1pt}
    \setlength{\parsep}{0pt}
    \setlength{\topsep}{2pt}
    \item MMEB training set: \url{https://huggingface.co/datasets/TIGER-Lab/MMEB-train}
    \item ViDoRe training set: \url{https://huggingface.co/datasets/vidore/colpali_train_set}
    \item MoCa hard negatives: \url{https://huggingface.co/datasets/moca-embed/MoCa-CL-Pairs}
    \item MMEB test set: \url{https://huggingface.co/datasets/TIGER-Lab/MMEB-eval}
    \item ViDoRe V1 test sets: \url{https://huggingface.co/collections/vidore/vidore-benchmark-667173f98e70a1c0fa4db00d}
    \item ViDoRe V2 test sets: \url{https://huggingface.co/collections/vidore/vidore-benchmark-v2-dev-67ae03e3924e85b36e7f53b0}
\end{itemize}
Model initialization uses the following public checkpoint:
\begin{itemize}
    \setlength{\itemsep}{1pt}
    \setlength{\parsep}{0pt}
    \setlength{\topsep}{2pt}
    \item ColQwen2.5-base: \url{https://huggingface.co/vidore/colqwen2.5-base}
\end{itemize}
\endgroup
Section~\ref{sec:experiments} defines the evaluation protocol, while Appendix~\ref{sec:appendix-setup} records the benchmark composition and complete training configuration.

\bibliographystyle{iclr2027_conference}
\bibliography{references}

@inproceedings{metaembed,
  title={{MetaEmbed}: Scaling Multimodal Retrieval at Test-Time with Flexible Late Interaction},
  author={Xiao, Zilin and Ma, Qi and Gu, Mengting and Chen, Chun-cheng Jason and Chen, Xintao and Ordonez, Vicente and Mohan, Vijai},
  booktitle={The Fourteenth International Conference on Learning Representations},
  year={2026},
  url={https://openreview.net/forum?id=yKDqg9HwZX}
}

@inproceedings{zhu2026mure,
  author    = {Fengbin Zhu and
               Zijing Cai and
               Yuzhe Wang and
               Pengyang Shao and
               Wenjie Wang and
               Fuli Feng and
               Richang Hong and
               Tat-Seng Chua},
  title     = {{MURE}: Hierarchical Multi-Resolution Encoding via
               Vision-Language Models for Visual Document Retrieval},
  booktitle = {Proceedings of the 2026 International Conference on
               Multimedia Retrieval},
  pages     = {69--78},
  publisher = {ACM},
  year      = {2026},
  doi       = {10.1145/3805622.3810864}
}

@inproceedings{radford2021learning,
  title        = {Learning Transferable Visual Models From Natural Language Supervision},
  author       = {Radford, Alec and Kim, Jong Wook and Hallacy, Chris and Ramesh, Aditya and Goh, Gabriel and Agarwal, Sandhini and Sastry, Girish and Askell, Amanda and Mishkin, Pamela and Clark, Jack and Krueger, Gretchen and Sutskever, Ilya},
  booktitle    = {Proceedings of the 38th International Conference on Machine Learning},
  pages        = {8748--8763},
  year         = {2021},
  volume       = {139},
  series       = {Proceedings of Machine Learning Research},
  publisher    = {PMLR},
  url          = {https://proceedings.mlr.press/v139/radford21a.html}
}

@inproceedings{zhai2023sigmoid,
  title        = {Sigmoid Loss for Language Image Pre-Training},
  author       = {Zhai, Xiaohua and Mustafa, Basil and Kolesnikov, Alexander and Beyer, Lucas},
  booktitle    = {Proceedings of the IEEE/CVF International Conference on Computer Vision},
  pages        = {11975--11986},
  year         = {2023},
  doi          = {10.1109/ICCV51070.2023.01100}
}

@article{beyer2024paligemma,
  title        = {{PaliGemma}: A Versatile 3B {VLM} for Transfer},
  author       = {Beyer, Lucas and Steiner, Andreas and Pinto, Andr{\'e} Susano and Kolesnikov, Alexander and Wang, Xiao and Salz, Daniel and Neumann, Maxim and Alabdulmohsin, Ibrahim and Tschannen, Michael and Bugliarello, Emanuele and Unterthiner, Thomas and Keysers, Daniel and Koppula, Skanda and Liu, Fangyu and Grycner, Adam and Gritsenko, Alexey and Houlsby, Neil and Kumar, Manoj and Rong, Keran and Eisenschlos, Julian and Kabra, Rishabh and Bauer, Matthias and Bo{\v{s}}njak, Matko and Chen, Xi and Minderer, Matthias and Voigtlaender, Paul and Bica, Ioana and Balazevic, Ivana and Puigcerver, Joan and Papalampidi, Pinelopi and Henaff, Olivier and Xiong, Xi and Soricut, Radu and Harmsen, Jeremiah and Zhai, Xiaohua},
  journal      = {arXiv preprint arXiv:2407.07726},
  year         = {2024},
  doi          = {10.48550/arXiv.2407.07726},
  url          = {https://arxiv.org/abs/2407.07726}
}

@inproceedings{khattab2020colbert,
  title={ColBERT: Efficient and Effective Passage Search via Contextualized Late Interaction over BERT},
  author={Khattab, Omar and Zaharia, Matei},
  booktitle={Proceedings of the 43rd International ACM SIGIR Conference on Research and Development in Information Retrieval},
  pages={39--48},
  year={2020}
}

@INPROCEEDINGS{visionzip2024,
  author={Yang, Senqiao and Chen, Yukang and Tian, Zhuotao and Wang, Chengyao and Li, Jingyao and Yu, Bei and Jia, Jiaya},
  booktitle={2025 IEEE/CVF Conference on Computer Vision and Pattern Recognition (CVPR)},
  title={VisionZip: Longer is Better but Not Necessary in Vision Language Models},
  year={2025},
  volume={},
  number={},
  pages={19792-19802},
  doi={10.1109/CVPR52734.2025.01843}}

@INPROCEEDINGS{folder2025,
  author={Wang, Haicheng and Yu, Zhemeng and Spadaro, Gabriele and Ju, Chen and Quétu, Victor and Xiao, Shuai and Tartaglione, Enzo},
  booktitle={2025 IEEE/CVF International Conference on Computer Vision (ICCV)},
  title={FOLDER: Accelerating Multi-Modal Large Language Models with Enhanced Performance},
  year={2025},
  volume={},
  number={},
  pages={23614-23625},
  doi={10.1109/ICCV51701.2025.02192}}

@inproceedings{dart2025,
  title={Stop looking for “important tokens” in multimodal language models: Duplication matters more},
  author={Wen, Zichen and Gao, Yifeng and Wang, Shaobo and Zhang, Junyuan and Zhang, Qintong and Li, Weijia and He, Conghui and Zhang, Linfeng},
  booktitle={Proceedings of the 2025 Conference on Empirical Methods in Natural Language Processing},
  pages={9972--9991},
  year={2025}
}

@inproceedings{jiang2025vlmvec,
  title={{VLM}2Vec: Training Vision-Language Models for Massive Multimodal Embedding Tasks},
  author={Ziyan Jiang and Rui Meng and Xinyi Yang and Semih Yavuz and Yingbo Zhou and Wenhu Chen},
  booktitle={The Thirteenth International Conference on Learning Representations},
  year={2025},
}

@inproceedings{ColPali,
  author       = {Manuel Faysse and
                  Hugues Sibille and
                  Tony Wu and
                  Bilel Omrani and
                  Gautier Viaud and
                  C{\'{e}}line Hudelot and
                  Pierre Colombo},
  title        = {ColPali: Efficient Document Retrieval with Vision Language Models},
  booktitle    = {The Thirteenth International Conference on Learning Representations,
                  {ICLR} 2025, Singapore, April 24-28, 2025},
  publisher    = {OpenReview.net},
  year         = {2025},
  bibsource    = {dblp computer science bibliography, https://dblp.org}
}

@InProceedings{refcoco,
author="Yu, Licheng
and Poirson, Patrick
and Yang, Shan
and Berg, Alexander C.
and Berg, Tamara L.",
editor="Leibe, Bastian
and Matas, Jiri
and Sebe, Nicu
and Welling, Max",
title="Modeling Context in Referring Expressions",
booktitle="Computer Vision -- ECCV 2016",
year="2016",
publisher="Springer International Publishing",
address="Cham",
pages="69--85",
isbn="978-3-319-46475-6"
}

@article{mace2025vidore,
  title={ViDoRe Benchmark V2: Raising the Bar for Visual Retrieval},
  author={Mac{\'e}, Quentin and Loison, Ant{\'o}nio and Faysse, Manuel},
  journal={arXiv preprint arXiv:2505.17166},
  year={2025}
}

@article{Qwen25-VL,
  author       = {Shuai Bai and
                  Keqin Chen and
                  Xuejing Liu and
                  Jialin Wang and
                  Wenbin Ge and
                  Sibo Song and
                  Kai Dang and
                  Peng Wang and
                  Shijie Wang and
                  Jun Tang and
                  Humen Zhong and
                  Yuanzhi Zhu and
                  Mingkun Yang and
                  Zhaohai Li and
                  Jianqiang Wan and
                  Pengfei Wang and
                  Wei Ding and
                  Zheren Fu and
                  Yiheng Xu and
                  Jiabo Ye and
                  Xi Zhang and
                  Tianbao Xie and
                  Zesen Cheng and
                  Hang Zhang and
                  Zhibo Yang and
                  Haiyang Xu and
                  Junyang Lin},
  title        = {{Qwen2.5-VL} Technical Report},
  journal      = {CoRR},
  volume       = {abs/2502.13923},
  year         = {2025},
  doi          = {10.48550/ARXIV.2502.13923},
  eprinttype    = {arXiv},
  eprint       = {2502.13923},
  bibsource    = {dblp computer science bibliography, https://dblp.org}
}

@article{kusupati2022matryoshka,
  title={Matryoshka representation learning},
  author={Kusupati, Aditya and Bhatt, Gantavya and Rege, Aniket and Wallingford, Matthew and Sinha, Aditya and Ramanujan, Vivek and Howard-Snyder, William and Chen, Kaifeng and Kakade, Sham and Jain, Prateek and others},
  journal={Advances in Neural Information Processing Systems},
  volume={35},
  pages={30233--30249},
  year={2022}
}

@inproceedings{cha2024honeybee,
  title={Honeybee: Locality-Enhanced Projector for Multimodal {LLM}},
  author={Cha, Junbum and Kang, Wooyoung and Mun, Jonghwan and Roh, Byungseok},
  booktitle={Proceedings of the IEEE/CVF Conference on Computer Vision and Pattern Recognition},
  pages={13817--13827},
  year={2024}
}

@article{liu2024visual,
  title={Visual Anchors Are Strong Information Aggregators for Multimodal Large Language Model},
  author={Liu, Haogeng and You, Quanzeng and Han, Xiaotian and Liu, Yongfei and Huang, Huaibo and He, Ran and Yang, Hongxia},
  journal={Advances in Neural Information Processing Systems},
  volume={37},
  pages={17696--17718},
  year={2024}
}

@inproceedings{gordo2016deep,
  title={Deep Image Retrieval: Learning Global Representations for Image Search},
  author={Gordo, Albert and Almaz{\'a}n, Jon and Revaud, Jerome and Larlus, Diane},
  booktitle={Computer Vision -- ECCV 2016},
  pages={241--257},
  year={2016},
  publisher={Springer},
  doi={10.1007/978-3-319-46466-4_15}
}

@inproceedings{deng2009imagenet,
  title={ImageNet: A Large-Scale Hierarchical Image Database},
  author={Deng, Jia and Dong, Wei and Socher, Richard and Li, Li-Jia and Li, Kai and Fei-Fei, Li},
  booktitle={2009 IEEE Conference on Computer Vision and Pattern Recognition},
  pages={248--255},
  year={2009},
  publisher={IEEE},
  doi={10.1109/CVPR.2009.5206848}
}

@article{zhang2024longclip,
  title={{Long-CLIP}: Unlocking the Long-Text Capability of {CLIP}},
  author={Zhang, Beichen and Zhang, Pan and Dong, Xiaoyi and Zang, Yuhang and Wang, Jiaqi},
  journal={arXiv preprint arXiv:2403.15378},
  year={2024},
  doi={10.48550/arXiv.2403.15378},
  url={https://arxiv.org/abs/2403.15378}
}

@inproceedings{hu2018learning,
  title={Learning Answer Embeddings for Visual Question Answering},
  author={Hu, Hexiang and Chao, Wei-Lun and Sha, Fei},
  booktitle={Proceedings of the IEEE Conference on Computer Vision and Pattern Recognition},
  pages={5428--5436},
  year={2018},
  doi={10.1109/CVPR.2018.00569}
}

@inproceedings{yao2022filip,
  title={{FILIP}: Fine-Grained Interactive Language-Image Pre-Training},
  author={Yao, Lewei and Huang, Runhui and Hou, Lu and Lu, Guansong and Niu, Minzhe and Xu, Hang and Liang, Xiaodan and Li, Zhenguo and Jiang, Xin and Xu, Chunjing},
  booktitle={International Conference on Learning Representations},
  year={2022},
  url={https://openreview.net/forum?id=cpDhcsEDC2}
}

@inproceedings{thrush2022winoground,
  title={Winoground: Probing Vision and Language Models for Visio-Linguistic Compositionality},
  author={Thrush, Tristan and Jiang, Ryan and Bartolo, Max and Singh, Amanpreet and Williams, Adina and Kiela, Douwe and Ross, Candace},
  booktitle={Proceedings of the IEEE/CVF Conference on Computer Vision and Pattern Recognition},
  pages={5238--5248},
  year={2022},
  doi={10.1109/CVPR52688.2022.00517}
}

@inproceedings{li2022blip,
  title={{BLIP}: Bootstrapping Language-Image Pre-Training for Unified Vision-Language Understanding and Generation},
  author={Li, Junnan and Li, Dongxu and Xiong, Caiming and Hoi, Steven C. H.},
  booktitle={Proceedings of the 39th International Conference on Machine Learning},
  volume={162},
  pages={12888--12900},
  year={2022},
  series={Proceedings of Machine Learning Research},
  publisher={PMLR},
  url={https://proceedings.mlr.press/v162/li22n.html}
}

@article{yu2022coca,
  title={{CoCa}: Contrastive Captioners Are Image-Text Foundation Models},
  author={Yu, Jiahui and Wang, Zirui and Vasudevan, Vijay and Yeung, Legg and Seyedhosseini, Mojtaba and Wu, Yonghui},
  journal={Transactions on Machine Learning Research},
  year={2022},
  url={https://openreview.net/forum?id=Ee277P3AYC}
}

@article{jiang2024e5v,
  title={{E5-V}: Universal Embeddings with Multimodal Large Language Models},
  author={Jiang, Ting and Song, Minghui and Zhang, Zihan and Huang, Haizhen and Deng, Weiwei and Sun, Feng and Zhang, Qi and Wang, Deqing and Zhuang, Fuzhen},
  journal={arXiv preprint arXiv:2407.12580},
  year={2024},
  doi={10.48550/arXiv.2407.12580},
  url={https://arxiv.org/abs/2407.12580}
}

@inproceedings{zhang2024gme,
  title={Bridging Modalities: Improving Universal Multimodal Retrieval by Multimodal Large Language Models},
  author={Zhang, Xin and Zhang, Yanzhao and Xie, Wen and Li, Mingxin and Dai, Ziqi and Long, Dingkun and Xie, Pengjun and Zhang, Meishan and Li, Wenjie and Zhang, Min},
  booktitle={Proceedings of the IEEE/CVF Conference on Computer Vision and Pattern Recognition},
  pages={9274--9285},
  year={2025}
}

@article{meng2025vlm2vecv2,
  title={{VLM2Vec-V2}: Advancing Multimodal Embedding for Videos, Images, and Visual Documents},
  author={Meng, Rui and Jiang, Ziyan and Liu, Ye and Su, Mingyi and Yang, Xinyi and Fu, Yuepeng and Qin, Can and Thirukovalluru, Raghuveer and Zhang, Xuan and Chen, Zeyuan and Xu, Ran and Xiong, Caiming and Zhou, Yingbo and Chen, Wenhu and Yavuz, Semih},
  journal={Transactions on Machine Learning Research},
  year={2026},
  url={https://openreview.net/forum?id=TpU38jbKIJ}
}

@inproceedings{wei2024uniir,
  title={{UniIR}: Training and Benchmarking Universal Multimodal Information Retrievers},
  author={Wei, Cong and Chen, Yang and Chen, Haonan and Hu, Hexiang and Zhang, Ge and Fu, Jie and Ritter, Alan and Chen, Wenhu},
  booktitle={Computer Vision -- ECCV 2024},
  series={Lecture Notes in Computer Science},
  volume={15145},
  pages={387--404},
  year={2024},
  publisher={Springer},
  doi={10.1007/978-3-031-73021-4_23}
}

@inproceedings{zhang2024magiclens,
  title={MagicLens: Self-Supervised Image Retrieval with Open-Ended Instructions},
  author={Zhang, Kai and Luan, Yi and Hu, Hexiang and Lee, Kenton and Qiao, Siyuan and Chen, Wenhu and Su, Yu and Chang, Ming-Wei},
  booktitle={Proceedings of the 41st International Conference on Machine Learning},
  year={2024},
  url={https://openreview.net/forum?id=Zc22RDtsvP}
}

@inproceedings{zhou2025megapairs,
  title={MegaPairs: Massive Data Synthesis for Universal Multimodal Retrieval},
  author={Zhou, Junjie and Xiong, Yongping and Liu, Zheng and Liu, Ze and Xiao, Shitao and Wang, Yueze and Zhao, Bo and Zhang, Chen Jason and Lian, Defu},
  booktitle={Proceedings of the 63rd Annual Meeting of the Association for Computational Linguistics (Volume 1: Long Papers)},
  pages={19076--19095},
  year={2025},
  publisher={Association for Computational Linguistics},
  doi={10.18653/v1/2025.acl-long.935},
  url={https://aclanthology.org/2025.acl-long.935/}
}

@inproceedings{chen2025mme5,
  title={mm{E}5: Improving Multimodal Multilingual Embeddings via High-quality Synthetic Data},
  author={Chen, Haonan and Wang, Liang and Yang, Nan and Zhu, Yutao and Zhao, Ziliang and Wei, Furu and Dou, Zhicheng},
  booktitle={Findings of the Association for Computational Linguistics: ACL 2025},
  pages={8254--8275},
  year={2025},
  publisher={Association for Computational Linguistics},
  doi={10.18653/v1/2025.findings-acl.433},
  url={https://aclanthology.org/2025.findings-acl.433/}
}

@inproceedings{visrag,
  title={VisRAG: Vision-Based Retrieval-Augmented Generation on Multi-Modality Documents},
  author={Yu, Shi and Tang, Chaoyue and Xu, Bokai and Cui, Junbo and Ran, Junhao and Yan, Yukun and Liu, Zhenghao and Wang, Shuo and Han, Xu and Liu, Zhiyuan and Sun, Maosong},
  booktitle={The Thirteenth International Conference on Learning Representations},
  year={2025},
  url={https://openreview.net/forum?id=zG459X3Xge}
}

@inproceedings{masry-etal-2025-colmate,
  title={ColMate: Contrastive Late Interaction and Masked Text for Multimodal Document Retrieval},
  author={Masry, Ahmed and Thakkar, Megh and Bechard, Patrice and Madhusudhan, Sathwik Tejaswi and Awal, Rabiul and Mishra, Shambhavi and Suresh, Akshay Kalkunte and Daruru, Srivatsava and Hoque, Enamul and Gella, Spandana and Scholak, Torsten and Rajeswar, Sai},
  booktitle={Proceedings of the 2025 Conference on Empirical Methods in Natural Language Processing: Industry Track},
  pages={2071--2080},
  year={2025},
  publisher={Association for Computational Linguistics},
  doi={10.18653/v1/2025.emnlp-industry.145},
  url={https://aclanthology.org/2025.emnlp-industry.145/}
}

@article{chen2024far,
  title={How Far Are We to {GPT-4V}? Closing the Gap to Commercial Multimodal Models with Open-Source Suites},
  author={Chen, Zhe and Wang, Weiyun and Tian, Hao and Ye, Shenglong and Gao, Zhangwei and Cui, Erfei and Tong, Wenwen and Hu, Kongzhi and Luo, Jiapeng and Ma, Zheng and others},
  journal={Science China Information Sciences},
  volume={67},
  number={12},
  pages={220101},
  year={2024},
  publisher={Springer}
}

@article{li2025tokenpacker,
  title={{TokenPacker}: Efficient Visual Projector for Multimodal {LLM}},
  author={Li, Wentong and Yuan, Yuqian and Liu, Jian and Tang, Dongqi and Wang, Song and Qin, Jie and Zhu, Jianke and Zhang, Lei},
  journal={International Journal of Computer Vision},
  volume={133},
  number={10},
  pages={6794--6812},
  year={2025},
  publisher={Springer}
}

@inproceedings{zhang2026llava,
  title={{LLaVA-UHD v2}: Exploiting Hierarchical Vision Granularity in {MLLMs} via Inverse Semantic Pyramid},
  author={Zhang, Yipeng and Liu, Yifan and Guo, Zonghao and Zhang, Yidan and Yang, Xuesong and Zhang, Xiaoying and Chen, Chi and Song, Jun and Yao, Yuan and Chua, Tat-Seng and others},
  booktitle={Proceedings of the AAAI Conference on Artificial Intelligence},
  volume={40},
  pages={12934--12942},
  year={2026}
}

@article{deng2026scope,
  title={{SCOPE}: Saliency-Coverage Oriented Token Pruning for Efficient Multimodal {LLMs}},
  author={Deng, Jinhong and Li, Wen and Zhou, Joey Tianyi and He, Yang},
  journal={Advances in Neural Information Processing Systems},
  volume={38},
  pages={161527--161552},
  year={2026}
}

@inproceedings{ma2025towards,
  title={Towards Storage-Efficient Visual Document Retrieval: An Empirical Study on Reducing Patch-Level Embeddings},
  author={Ma, Yubo and Li, Jinsong and Zang, Yuhang and Wu, Xiaobao and Dong, Xiaoyi and Zhang, Pan and Cao, Yuhang and Duan, Haodong and Wang, Jiaqi and Cao, Yixin and others},
  booktitle={Findings of the Association for Computational Linguistics: ACL 2025},
  pages={19568--19580},
  year={2025}
}

@inproceedings{qin2026multi,
author = {Qin, Hanxiang and Martin, Alexander and Jha, Rohan and Zuo, Chunsheng and Kriz, Reno and Van Durme, Benjamin},
title = {Multi-Vector Index Compression in Any Modality},
year = {2026},
isbn = {9798400725999},
publisher = {Association for Computing Machinery},
address = {New York, NY, USA},
url = {https://doi.org/10.1145/3805712.3809589},
doi = {10.1145/3805712.3809589},
booktitle = {Proceedings of the 49th International ACM SIGIR Conference on Research and Development in Information Retrieval},
pages = {1519–1530},
numpages = {12},
location = {Australia},
series = {SIGIR '26}
}

@article{park2026all,
  title={Do All Visual Tokens Matter Equally? Object-Evidence Preserving Token Merging for Vision-Language Retrieval},
  author={Park, Suhyeong and Jung, Junha and Park, Jungwoo and Kang, Jaewoo},
  journal={arXiv preprint arXiv:2607.04605},
  year={2026}
}

@article{clavie2024reducing,
  title={Reducing the footprint of multi-vector retrieval with minimal performance impact via token pooling},
  author={Clavi{\'e}, Benjamin and Chaffin, Antoine and Adams, Griffin},
  journal={arXiv preprint arXiv:2409.14683},
  year={2024}
}

@inproceedings{bolya2023tome,
  title={Token Merging: Your {ViT} But Faster},
  author={Bolya, Daniel and Fu, Cheng-Yang and Dai, Xiaoliang and Zhang, Peizhao and Feichtenhofer, Christoph and Hoffman, Judy},
  booktitle={The Eleventh International Conference on Learning Representations},
  year={2023},
  url={https://openreview.net/forum?id=JroZRaRw7Eu}
}

@inproceedings{prillo2020softsort,
  title={{SoftSort}: A Continuous Relaxation for the Argsort Operator},
  author={Prillo, Sebastian and Eisenschlos, Julian},
  booktitle={Proceedings of the 37th International Conference on Machine Learning},
  pages={7793--7802},
  volume={119},
  series={Proceedings of Machine Learning Research},
  publisher={PMLR},
  year={2020},
  url={https://proceedings.mlr.press/v119/prillo20a.html}
}

@article{wang2023onepeace,
  title={{ONE-PEACE}: Exploring One General Representation Model Toward Unlimited Modalities},
  author={Wang, Peng and Wang, Shijie and Lin, Junyang and Bai, Shuai and Zhou, Xiaohuan and Zhou, Jingren and Wang, Xinggang and Zhou, Chang},
  journal={arXiv preprint arXiv:2305.11172},
  year={2023},
  doi={10.48550/arXiv.2305.11172},
  url={https://arxiv.org/abs/2305.11172}
}

@inproceedings{ma2024dse,
  title={Unifying Multimodal Retrieval via Document Screenshot Embedding},
  author={Ma, Xueguang and Lin, Sheng-Chieh and Li, Minghan and Chen, Wenhu and Lin, Jimmy},
  booktitle={Proceedings of the 2024 Conference on Empirical Methods in Natural Language Processing},
  pages={6492--6505},
  year={2024},
  publisher={Association for Computational Linguistics},
  doi={10.18653/v1/2024.emnlp-main.373},
  url={https://aclanthology.org/2024.emnlp-main.373/}
}

@inproceedings{liu2025lamra,
  title={{LamRA}: Large Multimodal Model as Your Advanced Retrieval Assistant},
  author={Liu, Yikun and Zhang, Yajie and Cai, Jiayin and Jiang, Xiaolong and Hu, Yao and Yao, Jiangchao and Wang, Yanfeng and Xie, Weidi},
  booktitle={Proceedings of the IEEE/CVF Conference on Computer Vision and Pattern Recognition},
  pages={4015--4025},
  year={2025}
}

\clearpage
\appendix
\section{Experimental Details}
\label{sec:appendix-setup}

\paragraph{MMEB.}
The Massive Multimodal Embedding Benchmark (MMEB)~\citep{jiang2025vlmvec} evaluates general multimodal embedding across 36 tasks: 10 classification, 10 visual question answering, 12 retrieval, and 4 visual grounding tasks. Precision@1 is computed for each task and macro-averaged; the in-domain and out-of-domain partitions contain 20 and 16 tasks, respectively.

\paragraph{ViDoRe V1.}
ViDoRe V1~\citep{ColPali} contains 10 page-level visual document retrieval subsets spanning academic and practical domains, document types including text, tables, figures, and infographics, and both English and French content. We report NDCG@5 on each subset and their macro-average.

\paragraph{ViDoRe V2.}
ViDoRe V2~\citep{mace2025vidore} contains 7 more challenging out-of-domain subsets drawn from ESG, economics, and biomedical documents, including human-authored, synthetic, and multilingual settings. We report NDCG@5 on each subset and their macro-average.

\paragraph{Optimization Details.}
We optimize with AdamW using random seed 42. All models are trained for 30,000 steps with a global batch size of 64, using eight examples per GPU across eight NVIDIA A100 GPUs. The LoRA rank and scaling factor are both 32. The learning rate decays linearly from $10^{-4}$ to zero, and the LoRA dropout is 0.1. The contrastive temperature is 0.03. Training uses bfloat16, FlashAttention-2, and gradient checkpointing. All active nested prefixes receive supervision; an ablation changes this objective only when explicitly stated.

\section{Gradient Flow through Residual Homogeneity Compression}
\label{sec:appendix-rhc-gradient}

RHC combines discrete matching with differentiable value aggregation. Because argsort has zero input gradients almost everywhere~\citep{prillo2020softsort}, the matching indices are treated as fixed during backpropagation and recomputed after each update. As in hard token merging with average pooling~\citep{bolya2023tome}, gradients pass through the merged values; Equation~\ref{eq:training-objective} further connects the assessor to the retrieval loss without a straight-through estimator or auxiliary compression loss.

\paragraph{Fixed-Assignment Formulation.}
At a compressed stage $k$, let $\{\mathbf{g}_{k,i}\}_{i=1}^{M_k}$ be the projected tokens and $\mathbf{h}_i$ their contextualized features. Define the raw importance and gate logits as $\zeta_i=f_{\mathrm{imp}}(\mathbf{h}_i)$ and $u_i=f_{\mathrm{gate}}(\mathbf{h}_i)$. With equal weighting of importance and novelty, the relative importance $\mathcal{S}_i$, cross-stage novelty $\mathcal{N}_i$, gate $\mathcal{G}_i$, and modulation weight $\mathcal{W}_i$ give
\begin{equation}
  \mathcal{S}_i=\operatorname{MinMax}(\zeta)_i,\quad
  \mathcal{G}_i=\operatorname{sigmoid}(u_i),\quad
  \mathcal{W}_i=\tfrac{1}{2}(\mathcal{S}_i+\mathcal{N}_i),\quad
  \widetilde{\mathbf{g}}_{k,i}
  =(1+\rho\mathcal{G}_i\mathcal{W}_i)\mathbf{g}_{k,i}.
  \label{eq:rhc-stage-modulation}
\end{equation}
MinMax is computed across the stage, so $\mathcal{S}_i$ is relative rather than absolute. Here $\rho\geq0$ is the modulation strength. Since $\mathcal{S}_i,\mathcal{N}_i,\mathcal{G}_i\in[0,1]$, the modulation is bounded by $[1,1+\rho]$ and preserves each token's direction.

Let $R_k=|\mathbf{r}_k|$ be the number of retained tokens and $\boldsymbol{\Pi}^{(k)}\in\{0,1\}^{R_k\times M_k}$ the final partition induced by the recursive merge rounds, where $\Pi^{(k)}_{\ell i}=1$ assigns input $i$ to group $\ell$. With group mass $m_\ell=\sum_i\Pi^{(k)}_{\ell i}$ and pre-normalized mean $\mathbf{y}_\ell$, the aggregation is
\begin{equation}
  \mathbf{y}_\ell
  =\frac{1}{m_\ell}\sum_i
  \Pi^{(k)}_{\ell i}\widetilde{\mathbf{g}}_{k,i},
  \qquad
  \mathbf{r}_{k,\ell}
  =\frac{\mathbf{y}_\ell}{\lVert\mathbf{y}_\ell\rVert_2}.
  \label{eq:rhc-group-mean}
\end{equation}
This identity follows inductively because merging groups of masses $m_a$ and $m_b$ replaces their means by $(m_a\mathbf{y}_a+m_b\mathbf{y}_b)/(m_a+m_b)$, preserving the mean of all original members.

\paragraph{Almost-Everywhere Differentiability.}
Assume strict destination-selection and source-ranking margins at every merge round, unique MinMax extrema with positive range, unique maximizing coarse anchors where novelty is evaluated, and nonzero group means $\mathbf{y}_\ell$. Then $\boldsymbol{\Pi}^{(k)}$ is locally constant, and Equations~\ref{eq:rhc-stage-modulation}--\ref{eq:rhc-group-mean} define a differentiable value path conditioned on that partition. The chain rule therefore yields its exact local gradient. Switching ties are measure-zero boundaries at which the hard map is not differentiable.

\paragraph{Task-Aligned Gradient to the Assessor.}
For query $b$ with positive document $d_b$ at level $k$, let $s_{bc}^{(k)}$ and $\pi_{bc}^{(k)}$ be the score and softmax probability of candidate document $c$, respectively. Let $\mathcal{I}_k$ be the active-query set, $\lambda_k$ its level weight, and $\tau$ the contrastive temperature. Equation~\ref{eq:training-objective} give
\begin{equation}
  \frac{\partial\mathcal{L}_{\mathrm{train}}}
  {\partial s_{bc}^{(k)}}
  =
  \frac{\lambda_k}{|\mathcal{I}_k|\tau}
  \left(\pi_{bc}^{(k)}-\mathbb{I}[c=d_b]\right),
  \label{eq:rhc-contrastive-direction}
\end{equation}
where $\mathbb{I}[\cdot]$ is the indicator function. Thus the positive candidate receives a negative score gradient and every negative a positive one. Away from ties, the current MaxSim and TopK matches route this signal to the stage outputs.

Let $\boldsymbol{\psi}_\ell=\nabla_{\mathbf{r}_{k,\ell}}\mathcal{L}_{\mathrm{train}}$ be this upstream gradient, let $\mathbf{I}$ be the identity in the $d$-dimensional retrieval space, and let $\kappa(i)$ be the group containing input $i$. Averaging and L2 normalization give
\begin{equation}
  \boldsymbol{\epsilon}_i
  =
  \nabla_{\widetilde{\mathbf{g}}_{k,i}}\mathcal{L}_{\mathrm{train}}
  =
  \frac{\mathbf{I}-\mathbf{r}_{k,\kappa(i)}
  \mathbf{r}_{k,\kappa(i)}^\top}
  {m_{\kappa(i)}\lVert\mathbf{y}_{\kappa(i)}\rVert_2}
  \boldsymbol{\psi}_{\kappa(i)}.
  \label{eq:rhc-value-gradient}
\end{equation}
Thus $\boldsymbol{\epsilon}_i$ connects every retained group member to the loss. Define $\chi_i=\mathbf{g}_{k,i}^\top\boldsymbol{\epsilon}_i$ as the sensitivity to token $i$'s scalar contribution. Then
\begin{equation}
  \frac{\partial\mathcal{L}_{\mathrm{train}}}{\partial\mathcal{G}_i}
  =\rho\mathcal{W}_i\chi_i,
  \qquad
  \frac{\partial\mathcal{L}_{\mathrm{train}}}{\partial\mathcal{S}_i}
  =\frac{\partial\mathcal{L}_{\mathrm{train}}}{\partial\mathcal{N}_i}
  =\frac{\rho}{2}\mathcal{G}_i\chi_i.
  \label{eq:rhc-modulation-gradient}
\end{equation}
Since $\mathcal{G}_i=\operatorname{sigmoid}(u_i)$, the gate logit receives
$\partial\mathcal{L}_{\mathrm{train}}/\partial u_i
=(\partial\mathcal{L}_{\mathrm{train}}/\partial\mathcal{G}_i)
\mathcal{G}_i(1-\mathcal{G}_i)$.
For $\rho>0$ (and $\mathcal{W}_i>0$ for the gate), $\chi_i<0$ favors larger relative saliency and gating, whereas $\chi_i>0$ favors smaller values. Let $\boldsymbol{\zeta}=(\zeta_1,\ldots,\zeta_{M_k})$ and let $\mathbf{1}$ be the all-ones vector. For any scalar shift $\xi$,
\begin{equation}
  \operatorname{MinMax}(\boldsymbol{\zeta}+\xi\mathbf{1})
  =\operatorname{MinMax}(\boldsymbol{\zeta}),
  \qquad \xi\in\mathbb{R},
  \label{eq:rhc-minmax-shift}
\end{equation}
so uniformly lowering every raw logit leaves $\mathcal{S}$ unchanged and cannot reduce the objective.

For the MinMax operation, let $i_{\min}=\arg\min_j \zeta_j$, $i_{\max}=\arg\max_j \zeta_j$, and $\Delta=\zeta_{i_{\max}}-\zeta_{i_{\min}}$. Its Jacobian is
\begin{equation}
  J^{\mathrm{mm}}_{ij}
  =\frac{\partial\mathcal{S}_i}{\partial \zeta_j}
  =\frac{\delta_{ij}-\delta_{i_{\min}j}}{\Delta}
  -\frac{(\zeta_i-\zeta_{i_{\min}})
  (\delta_{i_{\max}j}-\delta_{i_{\min}j})}{\Delta^2},
  \label{eq:rhc-minmax-jacobian}
\end{equation}
where $\delta_{ij}$ is the Kronecker delta. For the importance branch, the chain rule maps $\partial\mathcal{L}_{\mathrm{train}}/\partial\mathcal{S}_i$ through $J^{\mathrm{mm}}$ and then through $f_{\mathrm{imp}}$. Separately, the sigmoid relation above maps the gate gradient through $f_{\mathrm{gate}}$. Hence, both heads receive retrieval supervision whenever their corresponding local derivatives are nonzero.

The learned saliency is aligned with hard preservation. Write $\mathcal{U}_{ij}=\mathcal{C}_{ij}-\alpha(\mathcal{S}_i+\eta\mathcal{N}_i)$, where $\mathcal{C}_{ij}$ is pairwise similarity, $\alpha>0$ is the preservation coefficient, and $\eta\geq0$ weights novelty. Its direct dependence on source saliency satisfies
\begin{equation}
  \frac{\partial\mathcal{U}_{ij}}{\partial\mathcal{S}_i}
  =-\alpha<0,
  \qquad
  \arg\max_j\mathcal{U}_{ij}
  =\arg\max_j\mathcal{C}_{ij}.
  \label{eq:rhc-priority-direction}
\end{equation}
Equation~\ref{eq:rhc-priority-direction} states monotonicity of the hard rule, not a gradient through the assignment. Thus $\mathcal{S}_i$ is interpreted as task-aligned relative importance: increasing it protects source $i$ without changing its preferred destination. During backpropagation, the partition is fixed and then recomputed after the parameter update. Nonzero assessor updates require an active retrieval signal and within-group relative variation. Later-stage novelty treats accumulated coarse anchors as fixed references, while every earlier stage remains directly supervised by its nested objective. This analysis establishes trainability and directional alignment; the empirical contribution of these signals is evaluated in Figure~\ref{fig:main-ablation-plots}(a) and Tables~\ref{tab:rhc-component-ablation}--\ref{tab:rhc-component-ablation-v2}.

\section{Detailed RHC Scoring-Signal Results}
\label{sec:appendix-rhc-signals}

We retrain two controlled variants using the same initialization, stage budgets, and optimization as the complete model. The \emph{w/o Importance Score} variant removes $\mathcal{S}_i$ from the preservation priority, giving $\mathcal{P}_i=\eta\mathcal{N}_i$, whereas \emph{w/o Novelty Score} removes $\mathcal{N}_i$, giving $\mathcal{P}_i=\mathcal{S}_i$. Both variants retain the similarity score $\mathcal{C}_{ij}$, the merging procedure, nested supervision, and late-interaction scoring.

\begin{table}[htbp]
  \centering
  \caption{Ablation of RHC scoring signals on all ViDoRe V1 subsets (NDCG@5 \%).}
  \label{tab:rhc-component-ablation}
  \vspace{3pt}
  \appendixtablestyle
  \setlength{\tabcolsep}{2.6pt}
  \begin{tabular}{@{}lrrrrrrrrrrr@{}}
    \toprule
    \textbf{Variant} & \textbf{ArxivQ} & \textbf{DocQ} & \textbf{InfoQ} & \textbf{TabF} & \textbf{TATQ} & \textbf{Shift} & \textbf{AI} & \textbf{Energy} & \textbf{Gov.} & \textbf{Health} & \textbf{Avg.} \\
    \midrule
    \rowcolor{gray!15}
    \textbf{\textsc{ResComEmb}} & 88.8 & 61.7 & 94.2 & 95.2 & 80.4 & 90.4 & 99.6 & 96.6 & 97.9 & 99.3 & 90.4 \\
    \hspace{0.8em}w/o Importance Score & 87.3 & 63.0 & 93.6 & 94.6 & 80.8 & 88.9 & 99.5 & 96.5 & 97.5 & 99.1 & 90.1 \\
    \hspace{0.8em}w/o Novelty Score & 86.6 & 62.5 & 93.8 & 94.2 & 80.7 & 87.4 & 99.3 & 96.5 & 96.5 & 99.2 & 89.7 \\
    \bottomrule
  \end{tabular}
\end{table}

\begin{table}[htbp]
  \centering
  \caption{Ablation of RHC scoring signals on all ViDoRe V2 subsets (NDCG@5 \%). Syn, Mul, and Bio denote synthetic, multilingual, and biomedical data.}
  \label{tab:rhc-component-ablation-v2}
  \vspace{3pt}
  \appendixtablestyle
  \setlength{\tabcolsep}{2.5pt}
  \begin{tabular}{@{}>{\raggedright\arraybackslash}p{0.245\linewidth}
      >{\centering\arraybackslash}p{0.11\linewidth}
      >{\centering\arraybackslash}p{0.075\linewidth}
      >{\centering\arraybackslash}p{0.075\linewidth}
      >{\centering\arraybackslash}p{0.13\linewidth}
      >{\centering\arraybackslash}p{0.055\linewidth}
      >{\centering\arraybackslash}p{0.09\linewidth}
      >{\centering\arraybackslash}p{0.055\linewidth}
      >{\centering\arraybackslash}p{0.063\linewidth}@{}}
    \toprule
    \textbf{Variant} & \shortstack{\textbf{ESG}\\\textbf{Human}} & \shortstack{\textbf{Eco}\\\textbf{Mul}} & \shortstack{\textbf{Bio}\\\textbf{Mul}} & \shortstack{\textbf{ESG}\\\textbf{Syn-Mul}} & \textbf{Bio} & \shortstack{\textbf{ESG}\\\textbf{Syn}} & \textbf{Eco} & \textbf{Avg.} \\
    \midrule
    \rowcolor{gray!15}
    \textbf{\textsc{ResComEmb}} & 69.8 & 56.0 & 60.0 & 57.1 & 64.8 & 60.9 & 62.7 & 61.6 \\
    \mbox{\hspace{0.8em}w/o Importance Score} & 68.4 & 54.7 & 57.3 & 55.9 & 60.5 & 59.6 & 60.0 & 59.5 \\
    \mbox{\hspace{0.8em}w/o Novelty Score} & 67.5 & 51.2 & 56.8 & 48.8 & 62.3 & 57.3 & 62.7 & 58.1 \\
    \bottomrule
  \end{tabular}
\end{table}

The aggregate degradation on ViDoRe V2 is concentrated rather than uniform. Removing the novelty score costs 4.8 points on Eco Mul and 8.3 points on ESG Syn-Mul, while removing the importance score has its largest effect of 4.3 points on Bio. Because RHC compresses coarse to fine, later stages without novelty rank candidates only by salience; this leaves cross-stage redundancy unpenalized, whereas novelty promotes candidates weakly covered by earlier stages. The concentration of the losses on multilingual and synthetic subsets suggests that this complementary coverage is more consequential under those distribution shifts.

\section{Detailed Late-Interaction Results}
\label{sec:appendix-interaction}

The directional weights used in the late-interaction score are computed from
the numbers of valid query and document tokens. Specifically,
\begin{equation}
\begin{aligned}
  w_1(q,d)
  &=\min\!\left\{\rho_{\max},
    \max\!\left(\frac{1}{2},\frac{N_d}{N_q+N_d}\right)\right\},\\
  w_2(q,d)&=1-w_1(q,d),
\end{aligned}
\label{eq:appendix-adaptive-weights}
\end{equation}
where $N_q$ and $N_d$ denote the numbers of valid query and document tokens,
respectively, and $\rho_{\max}$ caps the weight assigned to the
query-to-document direction. Thus, the two directions receive equal weight
when $N_q=N_d$, while longer documents increase the relative weight of
query-to-document support without eliminating the reverse direction.

\begin{table}[t]
  \centering
  \caption{Late-interaction ablations by benchmark and MMEB category. \emph{w/o Adaptive Weighting} keeps both directions with fixed $0.5/0.5$ weights; \emph{w/o Bidirectionality} retains only query-to-document TopK-mean MaxSim. ViDoRe reports NDCG@5 (\%); MMEB reports Precision@1 (\%). ViDoRe Avg.\ weights V1/V2 by their 10/7 subsets.}
  \label{tab:late-interaction-ablation}
  \vspace{3pt}
  \appendixtablestyle
  \begin{tabular}{@{}>{\raggedright\arraybackslash}p{0.32\linewidth}@{}*{8}{>{\centering\arraybackslash}p{0.085\linewidth}@{}}}
    \toprule
    \multirow{2}{*}{\textbf{Variant}} & \multicolumn{3}{c}{\textbf{ViDoRe}} & \multicolumn{5}{c}{\textbf{MMEB}} \\
    \cmidrule(lr){2-4}\cmidrule(lr){5-9}
    & \textbf{V1} & \textbf{V2} & \textbf{Avg.} & \textbf{Cls.} & \textbf{VQA} & \textbf{Ret.} & \textbf{Gnd.} & \textbf{Avg.} \\
    \midrule
    \textcolor{gray!70!black}{Vanilla MaxSim} & \textcolor{gray!70!black}{90.0} & \textcolor{gray!70!black}{59.8} & \textcolor{gray!70!black}{77.6} & \textcolor{gray!70!black}{41.3} & \textcolor{gray!70!black}{15.1} & \textcolor{gray!70!black}{62.7} & \textcolor{gray!70!black}{71.8} & \textcolor{gray!70!black}{44.5} \\
    \midrule
    \rowcolor{gray!15}
    \textbf{\textsc{ResComEmb}} & \textbf{90.4} & \textbf{61.6} & \textbf{78.5} & \textbf{63.1} & 61.1 & \textbf{69.5} & 87.9 & \textbf{67.4} \\
    \hspace{0.8em}w/o Mean & 90.0 & 59.1 & 77.3 & 54.6 & 41.3 & 64.5 & \textbf{89.3} & 58.1 \\
    \hspace{0.8em}w/o TopK & 89.1 & 54.5 & 74.9 & 61.2 & 55.7 & 66.7 & 81.4 & 63.8 \\
    \hspace{0.8em}w/o Adaptive Weighting & 89.0 & 54.0 & 74.6 & 62.8 & \textbf{61.3} & 67.4 & \textbf{89.3} & 66.9 \\
    \hspace{0.8em}w/o Bidirectionality & 90.0 & 59.7 & 77.5 & 61.9 & 57.5 & 67.2 & 88.6 & 65.4 \\
    \bottomrule
  \end{tabular}
\end{table}

Table~\ref{tab:late-interaction-ablation} demonstrates that the reported scoring operations address complementary failure modes. Reverting to vanilla MaxSim reduces the MMEB average by 22.9 points, while removing mean aggregation alone incurs a 9.3-point drop, confirming the importance of normalized evidence aggregation for heterogeneous multimodal inputs. TopK selection and length-adaptive weighting are particularly important on ViDoRe V2: removing TopK and replacing the adaptive weights with a fixed $0.5/0.5$ average lower performance by 7.1 and 7.6 points, respectively. Removing bidirectionality lowers the MMEB average by 2.0 points, isolating the contribution of reverse document-to-query evidence while retaining query-to-document TopK-mean MaxSim.

\section{Preliminary Multi-Granularity Study}
\label{sec:appendix-oracle}

Following the combined-selector diagnostic of MURE~\citep{zhu2026mure}, we train and evaluate global ($g_1$), aspect-ratio-aware intermediate ($g_2$), and fine-grained ($g_3$) models separately. A per-query oracle then retains the best retrieval outcome using ground-truth relevance, estimating the potential of adaptive multi-resolution perception rather than a deployable routing method.

\begin{figure}[H]
  \centering
  \includegraphics[width=0.9\linewidth]{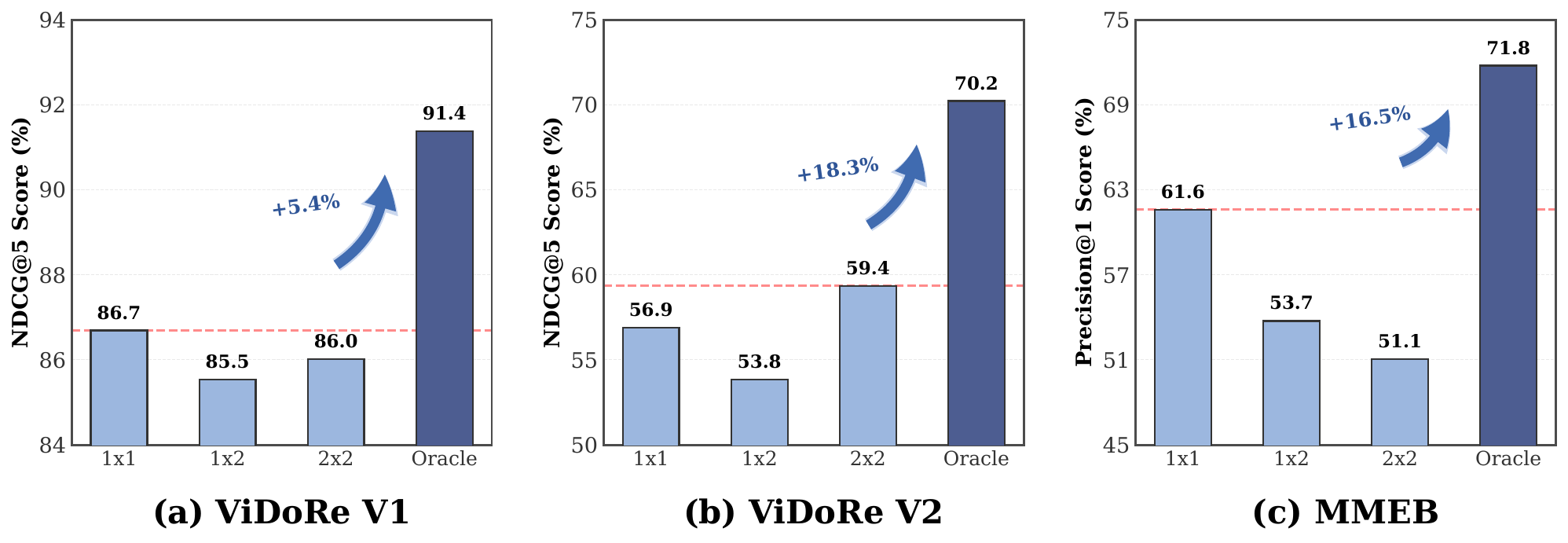}
  \caption{Single-granularity and oracle combined-selector results. ViDoRe V1/V2 use NDCG@5; MMEB uses Precision@1.}
  \label{fig:preliminary}
\end{figure}

The strongest fixed granularity differs across benchmarks, whereas the per-query oracle improves over it by 4.7 points on V1, 10.8 on V2, and 10.2 on MMEB Precision@1. The larger improvement on V2 indicates that complementary scale evidence becomes more important under domain shift, while the MMEB result extends this observation beyond document retrieval. These gains motivate preserving multiple visual scales before RHC compression.

\section{Additional Multi-Granularity Ablations}
\label{sec:appendix-multigranularity}

We remove one visual granularity at a time and compare each variant with the complete representation under the same total output budget of 384 visual tokens. The $w/o~g_1$, $w/o~g_2$, and $w/o~g_3$ variants use stage budgets $0/192/192$, $192/0/192$, and $192/192/0$, respectively, and supervise only the retained stages. All variants follow the same training setup. This controls representation capacity when measuring the contribution of each granularity.

\begin{table}[H]
  \centering
  \caption{Visual-granularity ablations at a fixed total budget of 384 visual tokens. ViDoRe reports NDCG@5 (\%); MMEB reports Precision@1 (\%). ViDoRe Avg.\ weights V1/V2 by their 10/7 subsets.}
  \label{tab:granularity-combination-ablation}
  \vspace{3pt}
  \appendixtablestyle
  \begin{tabular}{@{}>{\raggedright\arraybackslash}p{0.235\linewidth}@{}*{9}{>{\centering\arraybackslash}p{0.085\linewidth}@{}}}
    \toprule
    \multirow{2}{*}{\textbf{Variant}} & \multirow{2}{*}{\textbf{Tokens}} & \multicolumn{3}{c}{\textbf{ViDoRe}} & \multicolumn{5}{c}{\textbf{MMEB}} \\
    \cmidrule(lr){3-5} \cmidrule(l){6-10}
    & & \textbf{V1} & \textbf{V2} & \textbf{Avg.} & \textbf{Cls.} & \textbf{VQA} & \textbf{Ret.} & \textbf{Gnd.} & \textbf{Avg.} \\
    \midrule
    \rowcolor{gray!15}
    \textbf{\textsc{ResComEmb}} & 384 & \textbf{90.4} & 61.6 & 78.5 & \textbf{63.1} & 61.1 & \textbf{69.5} & 87.9 & \textbf{67.4} \\
    \hspace{0.8em}w/o $g_1$ & 384 & 90.0 & 59.1 & 77.3 & 62.1 & 61.2 & 66.4 & 89.5 & 66.3 \\
    \hspace{0.8em}w/o $g_2$ & 384 & \textbf{90.4} & \textbf{61.9} & \textbf{78.7} & 62.8 & \textbf{61.3} & 67.5 & \textbf{89.6} & 66.9 \\
    \hspace{0.8em}w/o $g_3$ & 384 & 90.2 & 61.5 & 78.4 & 62.9 & 61.2 & 67.8 & 89.2 & 67.0 \\
    \bottomrule
  \end{tabular}
\end{table}

Under the matched 384-vector budget, removing the global view causes the largest document-retrieval degradation, confirming its role as the structural anchor of the hierarchy. On MMEB, every two-view variant underperforms the complete representation, showing that global, intermediate, and fine-grained evidence are complementary for universal multimodal tasks. The full hierarchy is therefore selected for its strongest performance across benchmark families rather than for any single dataset.

\section{Additional Token-Budget Ablations}

Table~\ref{tab:appendix-token-budget-ablation} compares five symmetric per-granularity output token budgets.

\begin{table}[H]
  \centering
  \caption{Per-stage token-budget ablations for RHC. Stage budgets are written as $b_1/b_2/b_3$; Tokens counts the total visual budget. ViDoRe reports NDCG@5 (\%); MMEB reports Precision@1 (\%). ViDoRe Avg.\ weights V1/V2 by their 10/7 subsets.}
  \label{tab:appendix-token-budget-ablation}
  \vspace{3pt}
  \appendixtablestyle
  \begin{tabular*}{\linewidth}{@{\extracolsep{\fill}}lrrrrrrrrr@{}}
    \toprule
    \multirow{2}{*}{\textbf{Stage budget}} & \multirow{2}{*}{\textbf{Tokens}} & \multicolumn{3}{c}{\textbf{ViDoRe}} & \multicolumn{5}{c}{\textbf{MMEB}} \\
    \cmidrule(lr){3-5} \cmidrule(l){6-10}
    & & \textbf{V1} & \textbf{V2} & \textbf{Avg.} & \textbf{Cls.} & \textbf{VQA} & \textbf{Ret.} & \textbf{Gnd.} & \textbf{Avg.} \\
    \midrule
    $16/16/16$ & 48 & 86.2 & 55.5 & 73.6 & 62.0 & 60.9 & 66.7 & 89.6 & 66.3 \\
    $32/32/32$ & 96 & 88.9 & 58.6 & 76.4 & 62.3 & 61.3 & 67.7 & 89.8 & 66.9 \\
    $64/64/64$ & 192 & 89.5 & 60.6 & 77.6 & 62.6 & \textbf{61.4} & 68.6 & \textbf{90.2} & 67.3 \\
    $128/128/128$ & 384 & 90.4 & 61.6 & 78.5 & 63.1 & 61.1 & \textbf{69.5} & 87.9 & \textbf{67.4} \\
    $256/256/256$ & 768 & \textbf{90.6} & \textbf{61.7} & \textbf{78.7} & \textbf{63.2} & 61.1 & 66.9 & 88.5 & 66.7 \\
    \bottomrule
  \end{tabular*}
\end{table}

Most gains appear by 384 vectors: the 192-vector model nearly matches the main configuration, while 768 vectors add little to ViDoRe and lower the MMEB average. We therefore select $128/128/128$ as the operating point balancing capacity and storage. These budget variants are separately trained, unlike the nested prefixes in Table~\ref{tab:mrl-prefix-comparison}.

\end{document}